%% file: euclids_gif.tex
\documentclass[10pt,twocolumn,letterpaper]{article}

\usepackage[pagenumbers]{cvpr} 

\definecolor{cvprblue}{rgb}{0.21,0.49,0.74}
\usepackage[pagebackref,breaklinks,colorlinks,allcolors=cvprblue]{hyperref}

\usepackage{newfloat}
\usepackage{listings}
\usepackage{multirow} 

\usepackage[table]{xcolor}
\newcommand{\up}{\textcolor{green!60}{$\uparrow$}}

\definecolor{navyblue}{HTML}{0071BC}
\def\UP#1{\,\,\,\,\,#1\,\,\up}
\def\DOWN#1{#1}
\newcommand{\appref}[1]{Appendix~\ref{#1}}

\usepackage{tcolorbox} 
\newcommand{\hypbox}[2]{%
\begin{tcolorbox}[colback=white!98!black,colframe=white!30!black,boxsep=1.1pt,top=6.75pt]%
\vspace{1.75pt}%
\textbf{#1}\\[-0.575em]
\noindent\makebox[\textwidth]{\rule{\textwidth}{0.4pt}}
\\[0.25em]
#2
\end{tcolorbox}
}

\def\paperID{} 
\def\confName{}
\def\confYear{}

\title{Euclid's Gift: Enhancing Spatial Perception and Reasoning in Vision-Language Models via Geometric Surrogate Tasks}

\author{
    Shijie Lian\textsuperscript{\rm 1,\rm 2}\thanks{These authors contributed equally},  
    Changti Wu\textsuperscript{\rm 3,\rm 2}\footnotemark[1],  
    Laurence Tianruo Yang\textsuperscript{\rm 4,\rm 1}\thanks{Corresponding author}, 
    Hang Yuan\textsuperscript{\rm 2}\!\!,
    Bin Yu\textsuperscript{\rm 2}\!\!,
    Lei Zhang\textsuperscript{\rm 3}\!\!,
    Kai Chen\textsuperscript{\rm 5}\footnotemark[2] \\
    \textsuperscript{1}Huazhong University of Science and Technology  \space 
    \textsuperscript{2}Zhongguancun Academy \space
    \textsuperscript{3}East China \\ Normal University \space \textsuperscript{4}Zhengzhou University  \space \textsuperscript{5}Zhongguancun Institute of Artificial Intelligence \\
}

\begin{document}
\maketitle
\addtocontents{toc}{\protect\setcounter{tocdepth}{-1}}
\input{sec/0_abstract}

\input{sec/1_intro}
\input{sec/2_related}
\input{sec/3_method}
\input{sec/4_expeiment}
\input{sec/5_discussion}
\input{sec/6_conclusion}

{
    \small
    \bibliographystyle{ieeenat_fullname}
    \bibliography{euclids_gif}
}

\addtocontents{toc}{\protect\setcounter{tocdepth}{2}}
\input{sec/X_suppl}

\end{document}

%% file: sec/0_abstract.tex
\begin{figure*}[!t]
    \centering
    \vspace{-3mm}
    \includegraphics[width=1\linewidth]{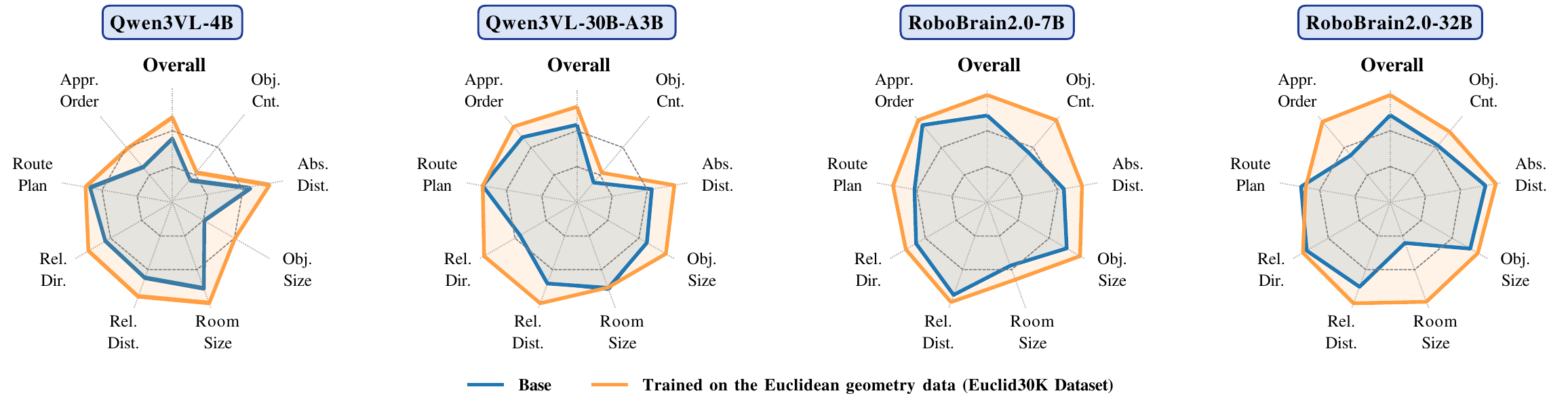}%
    \vspace{-3mm}
    \caption{Performance gains on VSIBench after model training on Euclid30K, for more complete data please refer to \cref{tab:vsibench}.}\label{fig:intro_show}
    \vspace{-3mm}
\end{figure*}

\begin{abstract}
    Spatial intelligence spans a rich suite of abilities, including visualising and transforming shapes, mentally rotating objects, judging relational positions and containment, and estimating numerosity.
    However, it still remains a critical unresolved challenge for Multimodal Large Language Models (MLLMs).
    To fill this gap, we propose to treat Euclidean geometry problem-solving as a surrogate task.
    Specifically, we meticulously constructed a curated multimodal dataset, called Euclid30K, comprising approximately 30K plane and solid geometry problems.
    Furthermore, to enable the model to learn and apply Euclidean principles from these geometry problems, we fine-tuned seven model variants (spanning 3--72B parameters) from the Qwen2.5VL, Qwen3VL, and RoboBrain2.0 families using Group Relative Policy Optimization (GRPO), inspiring the models to identify shapes, count, and relate entities, and perform multi-step deductive reasoning using Euclidean principles.
    Our experiments demonstrate that the resulting models achieve substantial zero-shot gains across four spatial reasoning benchmarks (Super-CLEVR, Omni3DBench, VSI-Bench, and MindCube) without any task-specific adaptations.
    Notably, after training on the Euclid30K, the mean VSI-Bench accuracy rose from 36.6\% to 41.8\% (+5.2\%), and the mean MindCube accuracy rose from 31.4\% to 38.1\% (+6.7\%).
    To our knowledge, this is the first systematic study showing that geometry-centric fine-tuning can confer vision-language models with broadly transferable spatial skills.
    Code and Euclid30K dataset can be found in \href{https://zgca-ai4edu.github.io/Euclids_Gift}{this}.
\end{abstract}

%% file: sec/1_intro.tex
\vspace{-4mm}

\section{Introduction}

\vspace{1mm}
\begin{quote}
\textit{“The whole is greater than the part.”}\\[-1em]
\begin{flushright}
\vspace{-3mm}
\textit{--- Euclid, \textit{Elements}~I, Common Notion 5}
\end{flushright}
\end{quote}

In recent years, multimodal large language models (MLLMs)~\cite{BLIP2_ICML_2023, Flamingo_NIPS_2022, LLaVa_2023_NIPS} have achieved remarkable success across a broad range of vision-language tasks, from image captioning and visual question answering (VQA) to document understanding~\cite{DeepSeek-VL_arXiv_2024, Gemini_1.5_arXiv_2024, Gemma3_arXiv_2025, Claude-4-Sonnet-20250514, KimiVL_arXiv_2025, Qwen2.5VL_2025_arXiv}.
State-of-the-art models like GPT-5~\cite{GPT5}, Gemini-2.5-Pro~\cite{Gemini_2.5_arXiv_2025}, and Qwen3VL-series~\cite{Qwen3VL_2025_arXiv} now rival or even surpass human performance on certain benchmarks, especially tasks requiring advanced language understanding or mathematical reasoning.
For instance, models like GPT-4o~\cite{GPT4o_2024_arXiv}, InternVL2-40B~\cite{Internvl_2024_CVPR}, and Qwen2.5VL-32B~\cite{Qwen2.5VL_2025_arXiv} have exceeded the average human score on the MathVista multimodal math benchmark~\cite{MathVista_2024_ICLR}, reflecting the rapid progress in integrating visual perception with high-level reasoning.

Despite recent progress, state-of-the-art MLLMs still fall short of genuine spatial intelligence~\cite{zhang2025call, VSIBench_2025_CVPR}.
Spatial intelligence involves perceiving and mentally manipulating spatial relationships and spans several tasks, such as estimating quantity, interpreting spatial relations, and understanding geometric configurations~\cite{gardner2011frames, VSIBench_2025_CVPR}.
Nowadays, leading vision-language models (VLMs) still make occasional mistakes on tasks that young children solve with ease, such as determining object orientation or identifying which object is the nearest neighbor to a given object on its left~\cite{Super-CLEVR_CVPR_2023, Omni3DBench_arXiv_2025}.
A recent evaluation on the Visual-Spatial Intelligence Benchmark (VSIBench) shows that more than 70\% of the recorded errors arise from faulty spatial reasoning, not from deficiencies in visual recognition or language parsing~\cite{VSIBench_2025_CVPR}.
This phenomenon is consistent with Moravec's paradox~\cite{moravec1988mind}, which suggests that high-level reasoning tasks are computationally simpler for VLM than low-level perceptual and sensorimotor skills.
Closing this gap is essential for the next generation of VLMs~\cite{dahou2025vision}.

Recent work on spatially aware VLMs, including Spatial-MLLM~\cite{Spatial-MLLM_2025_arXiv}, SpaceVLM~\cite{SpatialVLM_arXiv_2024}, VLM-3R~\cite{VLM-3R_arXiv_2025}, RoboBrain2.0~\cite{RoboBrain2.0_2025_TechnicalReport}, and SIMS-V~\cite{SIMS-V_2025_arXiv}, attempts to provide specially constructed spatial datasets to improve model performance. 
However, tasks in these spatial datasets usually cover only a subset of real-world spatial tasks and may not enhance the model's overall spatial intelligence.
For example, Spatial-MLLM collects data from ScanQA~\cite{ScanQA_2022_CVPR}, SQA3D~\cite{SQA3D_2022_ICLR}, and self-curated spatial QA data, and follows the eight tasks introduced in VSI-Bench~\cite{VSIBench_2025_CVPR} to build the Spatial-MLLM-120K dataset. 
The trained model therefore achieves state-of-the-art results on VSI-Bench, ScanQA, and SQA3D. However, its accuracy drops on the out-of-domain MindCube benchmark~\cite{MindCube_arXiv_2025}.
This highlights a critical challenge in the field: while fine-tuning on task-specific datasets can achieve high in-domain performance, it may lead to over-specialization and fail to cultivate a more fundamental, generalizable spatial intelligence.
To bridge this gap, VLMs must learn from a broader and more foundational range of spatial phenomena, thus extending their capabilities beyond the limitations of any single dataset.

In order to develop generalized spatial skills beyond any single benchmark, we attempt to explore a novel training paradigm that incorporates solving geometric problems as a surrogate task for enhancing spatial intelligence in the VLM.\@
Geometry compresses centuries of mathematical study into formal descriptions of space~\cite{Gray2013EpistemologyGeometry}, and to some extent it shapes our spatial representations in the world~\cite{Euclidean_2024_NatureCommunications}.
Therefore, learning to solve planar and solid geometry problems forces the model to internalize the axioms and constraints of Euclidean geometry, and provides the model with stronger out-of-domain generalization capabilities, because these principles are universal and independent of any single task.
As shown in \cref{fig:intro_show}, these low-level geometry priors, like “Euclid's Gift”, provide a principled foundation that supports zero-shot transfer to a wide range of downstream spatial tasks.

This suggests that the abilities required to solve geometric problems, including recognizing shapes and configurations, inferring spatial relationships such as parallel, angular, and relative positions, calculating or measuring geometric elements, and performing multistep logical reasoning, are also required for spatial perception tasks, like object counting, relational localization, and size estimation~\cite{Cheng_2024_CVPR, VSIBench_2025_CVPR}.
Thus, by using geometry as a surrogate task, we aim to instill foundational Euclidean priors that support a significant and critical subset of spatial intelligence, i.e., static spatial perception and reasoning~\cite{clements1992geometry, pittalis2010types}.

To enable geometry-centric training, we curated and released the Euclid30K dataset, containing 29,695 geometry problems from open-source datasets including Geometry3K~\cite{Geometry3K_AACL_2021}, MMK12~\cite{MMK12_arXiv_2025}, SolidGeo~\cite{SolidGeo_2025_arXiv}, and WeMath2~\cite{wemath2_arXiv_2025}, along with newly collected middle- and high-school exercises and competition problems.
Due to the existing corpus's significant bias toward plane geometry, we expanded Euclid30K with 3,996 solid geometry problems collected from textbooks and exercise materials.
Solid geometry exposes the model to intuitive 3D concepts such as perspective and occlusion, polyhedral structures, and spatial rotation, which are equally crucial for constructing robust spatial prior knowledge.
Additionally, all answers were reformatted for direct recognition by the rule-based reward function implemented via MathVerify~\cite{Math-Verify}. 
 
Furthermore, in order to attribute the performance improvement strictly to the supervision of geometric knowledge rather than the interference of complex algorithms or data augmentation, we intentionally used the well-established GRPO framework and Euclid30K to train the models.
Specifically, we used GRPO to fine-tune the Qwen2.5VL series (3B, 7B, 72B parameters), the Qwen3VL series (4B, 8B, 30B parameters), and the RoboBrain2.0 series (7B, 32B parameters). 
The resulting geometry-trained models deliver consistent gains on Super-CLEVR~\cite{Super-CLEVR_CVPR_2023}, Omni3D-Bench~\cite{Omni3DBench_arXiv_2025}, VSI-Bench~\cite{VSIBench_2025_CVPR}, and MindCube~\cite{MindCube_arXiv_2025}. 
This suggests that the abstract geometric knowledge distilled from Euclid30K can be migrated to different spatial tasks and supports the model's static spatial perception and reasoning capabilities, is an effective alternative for enhancing spatial intelligence in VLMs.

To our knowledge, this is the first work to use geometry problems as surrogate tasks to cultivate spatial intelligence in general-purpose VLMs. 
Unlike prior approaches that fine-tuned models on a single skill, our training paradigm uses geometric tasks to endow the model with a deeper, principle-driven understanding of space.
 This allows the model to go beyond the baseline in multiple spatial reasoning tasks without additional task-specific training.
In summary, our contributions are as follows:
\begin{itemize}
    \item We demonstrate that tackling geometry tasks can serve as an effective surrogate task for spatial intelligence.
          Learning to solve geometry problems assists models in acquiring the most basic perceptions of space, such as the axioms of Euclidean geometry. 
          These low-level priors provide a principled foundation that supports the zero-shot transfer to a wide range of downstream spatial tasks.

    \item We collected and constructed the Euclid30K, a VQA dataset of 29,695 geometry questions. 
          Euclid30K offers a diverse and challenging set of problems covering a wide range of geometric concepts, designed to help models learn formal descriptions of space from geometric tasks. 

    \item Extensive experiments on the Qwen2.5VL, Qwen3VL, and RoboBrain2.0 show significant performance gains on four spatial benchmarks (Super-CLEVR, Omni3D-Bench, VSI-Bench, and MindCube), providing empirical evidence that the geometric curriculum reliably enhances spatial reasoning across diverse evaluation settings.
\end{itemize}

%% file: sec/2_related.tex
\section{Related Work}

\subsection{Spatial Intelligence}
Spatial intelligence is the capacity to reason about spatial relations, including mental rotation, viewpoint switching, and object counting~\cite{gardner2011frames, VSIBench_2025_CVPR}.
Benchmarks such as CLEVR~\cite{Clevr_2017_CVPR}, Super-CLEVR~\cite{Super-CLEVR_CVPR_2023}, Omni3D-Bench~\cite{Omni3DBench_arXiv_2025}, VSI-Bench~\cite{VSIBench_2025_CVPR}, and MindCube~\cite{MindCube_arXiv_2025} show that accurate visual recognition alone is not enough; models must also perform explicit spatial reasoning.
Early VLMs, such as ViLT~\cite{vilt_2021_icml}, METER~\cite{meter_2022_cvpr}, Flamingo~\cite{Flamingo_NIPS_2022}, and PaLI~\cite{pali_2022_arxiv}, improved perception yet still struggled with counting and orientation.
Recent spatial-aware systems such as SpaceVLM~\cite{SpatialVLM_arXiv_2024}, Spatial-MLLM~\cite{Spatial-MLLM_2025_arXiv}, SpatialLLM~\cite{ma2025spatialllm}, RoboBrain~\cite{Robobrain_arXiv_2025}, and RoboBrain 2.0~\cite{RoboBrain2.0_2025_TechnicalReport} fine-tune on curated spatial datasets, but these datasets cover only part of real-world scenarios and are costly to build.
Our approach instead trains on readily available Euclidean geometry problems, treating them as surrogate tasks that transfer their principles to general visual-spatial perception and reasoning without costly data collection or architectural changes.

\subsection{Multimodal Math \& Geometry Datasets} 

Research on multimodal math reasoning has produced datasets with progressively broaden both scope and format. 
GeoQA~\cite{GeoQA_2021_ACL-IJCNLP}, Geometry3K~\cite{Geometry3K_AACL_2021}, and UniGeo~\cite{UniGeo_2022_EMNLP} pioneered the use of image-text pairs for plane-geometry question answering, laying the groundwork for visual-symbolic reasoning. 
Subsequent datasets, such as MMMU~\cite{MMMU_2024_CVPR}, MathVista~\cite{MathVista_2024_ICLR}, We-Math~\cite{We-math_2024_arXiv}, We-Math2~\cite{wemath2_arXiv_2025}, MMMU-Pro~\cite{mmmu-pro_2025_ACL}, and GeoSense~\cite{Geosense_2025_arXiv}, expanded to a wider array of math domains and introduced more challenging multimodal tasks. 
The recently released SolidGeo3K~\cite{SolidGeo_2025_arXiv} benchmark focuses on 3D geometry, providing skill tags and difficulty annotations to establish a dedicated testing platform for solid geometry reasoning.

%% file: sec/3_method.tex
\section{Method} 

This section first presents a domain-adaptation perspective that motivates using geometric problem solving to enhance spatial intelligence, then describes the construction of Euclid30K and our GRPO-based training framework.
We further elaborate on the connection between geometry and spatial cognition from an educational-psychology viewpoint in \appref{sec:edu-psych}.

\subsection{Motivation: A Domain-Adaptation Perspective}\label{ssec:motivation}

This subsection presents a domain-adaptation view that motivates why training on geometric problem solving serves as an effective surrogate for spatial intelligence.

Let the source distribution \(\mathcal{D}_S\) denote geometry problem solving (e.g., Euclid30K) and the target distribution \(\mathcal{D}_T\) denote spatial intelligence. We train a VLM policy \(\pi_{\theta}\) on the source. The policy induces a hypothesis:
\begin{equation}
 h(x) = \mathbb{I}\big[\operatorname{Ans}(x;\pi_{\theta}) \text{ is correct}\big] \in \{0,1\}.
\end{equation}
Let \(\mathcal{H}\) be the hypothesis class containing \(h\). For \(h,h'\in\mathcal{H}\), define the error (disagreement probabilities):
\[
 \epsilon_S(h,h') := \Pr_{x\sim\mathcal{D}_S}[h(x)\neq h'(x)],
\]
Specifically, when $f$ is an absolutely correct hypothesis, we abbreviate $\epsilon_S(h,f)$ as $\epsilon_S(h)$. The definition of $\epsilon_T(h,h')$ and $\epsilon_T(h)$ follows similarly.
Furthermore, the \(\mathcal{H}\Delta\mathcal{H}\) distance~\cite{BenDavid2010} between  \(\mathcal{D}_S\) and  \(\mathcal{D}_T\) is:
\begin{equation}
 d_{\mathcal{H}\Delta\mathcal{H}}(\mathcal{D}_S,\mathcal{D}_T)
 = 2\,\sup_{h,h'\in\mathcal{H}}\Big|
 \epsilon_S(h,h')-\epsilon_T(h,h')
 \Big|.
\end{equation}
Based on this definition, we can invoke the standard domain-adaptation bound~\cite{BenDavid2010}, which connects the $\epsilon_T(h)$ to the $\epsilon_S(h)$ through the $\mathcal{H}\Delta \mathcal{H}$ distance.

\hypbox{Standard Domain-Adaptation Bound}{For any \(h\in\mathcal{H}\), we have:
\begin{equation}
 \epsilon_T(h) \lesssim \epsilon_S(h) + \tfrac{1}{2}\, d_{\mathcal{H}\Delta\mathcal{H}}(\mathcal{D}_S,\mathcal{D}_T).
\label{eq:supp-da-bound}
\end{equation}
(A detailed proof can be found in \appref{sec:appendex_rationale}.
}

In light of the bound in \cref{eq:supp-da-bound}, the target error is governed by the source error $\epsilon_S(h)$ and the distribution discrepancy \(d_{\mathcal{H}\Delta\mathcal{H}}(\mathcal{D}_S,\mathcal{D}_T)\).
In practice, \(\epsilon_S(h)\) can typically be reduced through optimization and data curation, which makes the discrepancy term pivotal~\cite{BenDavid2010,mansour2009domain}.
Consequently, if $d_{\mathcal{H}\Delta\mathcal{H}}(\mathcal{D}_S,\mathcal{D}_T)$ is small, we can regard the source distribution $\mathcal{D}_S$ as a surrogate for the target distribution $\mathcal{D}_T$. 

We hypothesize that formal Euclidean geometry compresses a broad set of spatial regularities—congruence, similarity, perspective, parallelism, intersection, and positional reasoning—into theorem-like constraints reused across downstream spatial-intelligence tasks.
Compared to training on a narrow sub-skill (\textit{e.g.}, object counting, depth ordering, or size estimation), this breadth plausibly yields a smaller \(d_{\mathcal{H}\Delta\mathcal{H}}\) between geometry and target tasks, consistent with cognitive-science evidence on the generality of geometric knowledge in human perception and reasoning~\cite{lake2017building,feldman2003visual}.
We further elaborate on this connection from an educational-psychology perspective in \appref{sec:edu-psych}. 

Since the spatial-intelligence target-domain distribution is unknown and there is no canonical proxy dataset $\hat{\mathcal{D}}_T$ that faithfully spans the full space of spatial-intelligence tasks, $d_{\mathcal{H}\Delta\mathcal{H}}$ is not directly observable in practice.
Nevertheless, our subsequent empirical results align with this qualitative hypothesis: after Euclid30K fine-tuning, models exhibit consistent gains across Omni3D-Bench, VSI-Bench, Super-CLEVR, and MindCube (in Tables~\ref{tab:vsibench},\ref{tab:val_and_test}, and~\ref{tab:mindcube}), and qualitative cases in the Appendix (Figs.~\ref{fig:comp1}–\ref{fig:comp8}) further illustrate acquired, transferable skills such as perspective (near to far size), parallelism and similarity, and positional inference. 
These observations provide a principled rationale for geometry as a surrogate task for spatial intelligence.

\input{tab/euclid30k_statistics}

\subsection{Euclid30K}\label{ssec:dataset}

\begin{figure}[!t]
\centering
\includegraphics[width=0.95\linewidth]{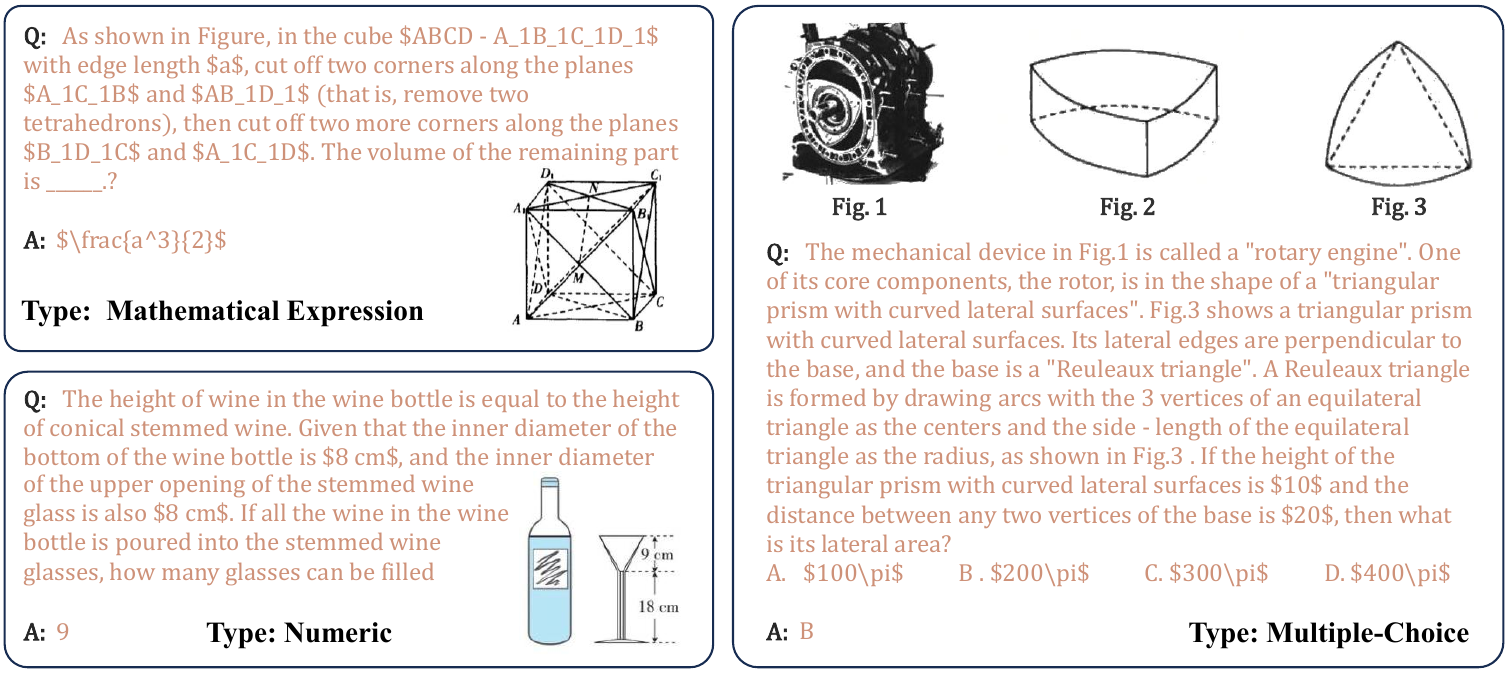}%
\caption{The examples of the newly collected questions in Euclid30K. More examples can be found in the appendix.}
\label{fig:data_show}
\vspace{-3mm}
\end{figure}

Plane and solid geometry give axiomatized abstractions of spatial phenomena. Training on such problems compels models to internalize Euclidean constraints such as angle and ratio preservation, similarity, and congruence, thereby providing an effective surrogate curriculum for cultivating spatial perception and reasoning skills.

Unfortunately, there are currently no large-scale, high-quality training datasets tailored for diverse geometric problems.
To address this, we filtered the required corpus from open-source resources such as Geometry3K~\cite{Geometry3K_AACL_2021}, MMK12~\cite{MMK12_arXiv_2025}, SolidGeo3K~\cite{SolidGeo_2025_arXiv}, and WeMath2~\cite{wemath2_arXiv_2025}. 
We employed Qwen2.5-VL-72B as an annotator to determine whether a problem belongs to plane or solid geometry.

After filtering the existing corpus, we identified a significant imbalance: only about 7,000 solid geometry problems remained, while there were about 20,000 plane geometry problems.
However, solid geometry encompasses more explicit three-dimensional spatial phenomena (\textit{e.g.}, perspective invariance, polyhedron truncation, volume-area relationships), which are equally crucial for VLMs in learning spatial knowledge.
Furthermore, existing solid geometry problems predominantly focus on shape recognition, coordinate/angle/area calculations, and similar question types, with insufficient coverage of richer problem types.
To address these gaps, we newly collected approximately 4,500 additional problems from K-12 textbooks and competition practice books, covering positional relationships, dynamic geometry, folding/unfolding, and contextualized word problems.
All newly collected problems are used with publisher permission for academic purposes.
\begin{figure*}[!t]
\centering
\includegraphics[width=1\linewidth]{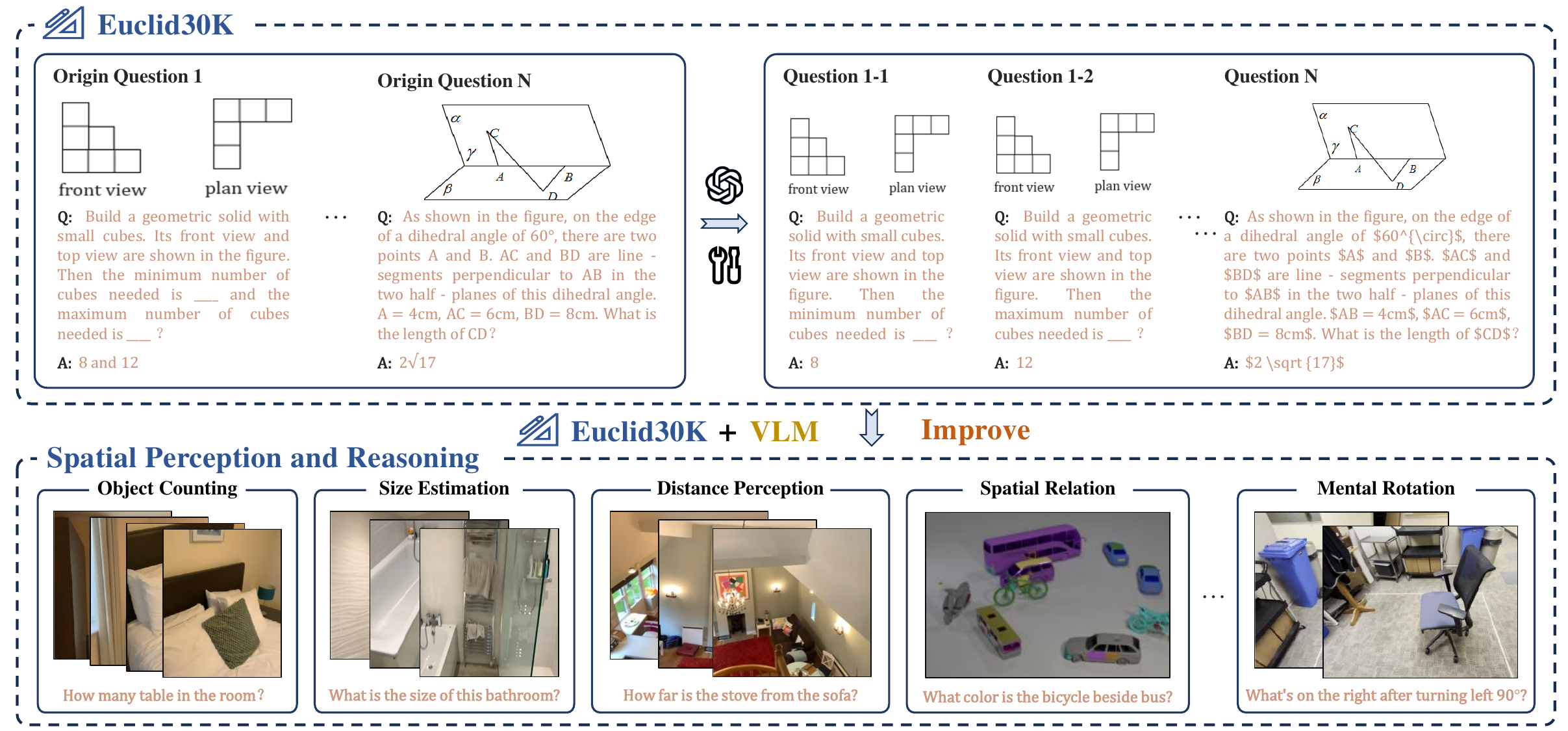}%
\caption{Enhancing spatial perception and reasoning capabilities in models using the geometric problem-solving dataset (Euclid30K).}\label{fig:main_fig}
\end{figure*}

As a result, we compiled about 32,500 candidate questions.
Then, to further ensure the quality of collected geometric data, we designed a three-stage filtering process:
\begin{itemize}
  \item \textbf{Duplicate Filtering:} Since even similar texts can lead to vastly different meanings or solution processes when paired with different images, we uniquely identify each question through its image. Specifically, we use image-based perceptual hashing to filter duplicate questions.
  \item \textbf{Question Splitting:} Many materials contain multiple sub-questions under a single main question. We utilize the GPT-4o API~\cite{GPT4o_2024_arXiv} to detect and enumerate each sub-question, splitting them into independent instances.
  \item \textbf{Formula Formatting:} We standardize formulae in questions and answers to LaTeX format via the DeepSeek-V3.1 API~\cite{deepseekai2024deepseekv3technicalreport}. This ensures answers can be correctly parsed by MathVerify~\cite{Math-Verify} for verification. For example, we replace~'$\pi$/2'~with~'\scalebox{0.7}{\textbackslash}frac\scalebox{0.7}{\{\textbackslash} pi\scalebox{0.7}{\}\{}2\scalebox{0.7}{\}}'. 
\end{itemize}

All conversions are subsequently verified and, when necessary, corrected by human annotators to ensure the final dataset is correct and consistent.
After the aforementioned filtering, splitting, formatting and human verification, we ultimately obtained 29,695 geometry problems, which were compiled into the Euclid30K dataset. 
Summary statistics about Euclid30K are reported in Table~\ref{tab:statistics}.
Additionally, we also list some examples of our newly collected data in \cref{fig:data_show}. 
Notably, the final Euclid30K dataset contains 3,996 newly collected/generated solid geometry problems, exceeding SolidGeo~\cite{SolidGeo_2025_arXiv}, the largest prior solid geometry dataset, which contained 3,113 problems.

Finally, each instance in Euclid30K is represented as a triple $<\text{image(s)}~i,~\text{text problem}~p,~\text{answer}~a>$, where $i$ may contain one or multiple figures, $p$ states the given conditions and questions, and $a$ is one of: a LaTeX expression, a numeric value, or the index of the correct option.

\subsection{RL Training in Euclid30K}\label{ssec:tuning}

We follow the standard GRPO training practice \citep{GRPO_DeepSeekMath_arXiv_2024} to enhance the geometry-solving ability of the Qwen2.5VL (3B, 72B),  Qwen3VL (4B, 8B, 30B), and the RoboBrain2.0 (7B, 32B), but introduce some modifications based on recent work~\cite{DAPO_arXiv_2025} to accelerate training.
Specifically, during training, we first sample a set of outputs $\{o_i\}_{i=1}^G$ for each question $q=\,\,<\emph{image\,\,i,\,\, textual problem\,\,p}>$ from the policy model $\pi_{\theta_{\text{old}}}$, and computing $\mathcal{J}(\theta)$: 
\begin{equation}\label{eq:loss1}
\mathcal{J}(\theta)=
\frac{1}{\gamma}\sum_{i=1}^{G}\sum_{t=1}^{|o_i|}
\min\bigl(
 r_{i,t}(\theta) \hat A_i,\,
\text{clip}( r_{i,t}(\theta), 1\pm\epsilon)\hat A_i
\bigr),
\end{equation}
where $\gamma=\sum_{i=1}^G|o_i|$ and $ r_{i,t}(\theta)=\frac{\pi_\theta(o_{i,t}|q,o_{i,<t})}{\pi_{\theta_{\text{old}}}(o_{i,t}|q,o_{i,<t})}$. Advantage function $\hat A_i = \frac{R_i - \text{mean}(\{R_i\}^G_{i=1})}{\text{std}(\{R_i\}^G_{i=1})}$ computed using the group-level rewards $\{R_i\}_{i=1}^G$.
In Eq.~\ref{eq:loss1}, we follow DAPO~\cite{DAPO_arXiv_2025} using the token-level policy gradient loss and ensure that each batch has valid gradients by over-sampling and filtering out prompts whose answers are either entirely correct or entirely wrong.
Then, we optimize the policy model~$\pi_{\theta}$ by maximizing the following objective:
\begin{equation}\label{eq:loss2}
\mathcal{L}_{\text{GRPO}}(\theta) = \ \mathbb{E}_{q,o_i}\left[
\mathcal{J}(\theta) - \beta \cdot \text{KL}[\pi_\theta \| \pi_{{\text{ref}}}]
\right].
\end{equation}

\input{tab/tab_vsibench}

In reinforcement learning with verifiable rewards for LLMs, the design of the reward function is critical~\cite{GRPO_DeepSeekMath_arXiv_2024, Spatial-MLLM_2025_arXiv}. In addition to a formatting reward applied to all task types, we introduce task-dependent reward modelling to ensure that it accurately reflects the proximity between the predicted and ground-truth answers. 
Specifically, if the answer is a mathematical expression containing variables in LaTeX format, we invoke MathVerify~\cite{Math-Verify} to perform exact symbolic equivalence checking, so algebraically identical forms (\textit{e.g.}, $2\pi r$ and $(2r)\pi$) receive the same reward. 
For purely numeric answers, to mitigate the risk of reward hacking~\cite{rewardhack_2024_blog}, we forgo the conventional mean relative accuracy (MRA) metric~\cite{VSIBench_2025_CVPR} and instead grant the reward if the prediction lies within ±1\% band around the ground truth:
\begin{equation}\label{eq:reward}
R \;=\;
\begin{cases}
1, & \displaystyle\left|\frac{pred - gt}{gt}\right| \le 0.01, \\[6pt]
0, & \text{otherwise},
\end{cases}
\end{equation}
where $pred$ is the model prediction and $gt$ is the ground truth value. 
This strategy filters out rough answers that have no practical relevance to the geometric conclusions, while allowing for rounding or floating-point truncation errors that are common in the generation process.
Further, to prevent unit mismatches from interfering with reward computation, we manually append unit prompts (\textit{e.g.}, Please answer in meters) to questions in Euclid30K whenever the ground-truth answer carries a specific unit.
In addition, for multiple choice questions, we use exact match reward.

%% file: tab/euclid30k_statistics.tex
\begin{table}[!t]
\centering
\renewcommand{\arraystretch}{1}
\setlength\tabcolsep{3pt} 
\resizebox{0.95\linewidth}{!}
{
  \begin{tabular}{lr}
    \toprule
    \textbf{Statistic} & \textbf{Number} \\
    \midrule
    \rowcolor{navyblue!10}\textbf{Total Number}& \textbf{29,695}\\
    \quad\enspace Mathematical Expression & 16,804 \\
    \quad\enspace Numeric & 6,321 \\
    \quad\enspace Multiple-Choice & 2,618  \\
    \midrule
    \rowcolor{navyblue!10}\textbf{Type}&\\
    \quad\enspace Plane (2D) / Solid (3D) & 18,577\,\,/ 11,118 \\  
    \midrule
    \rowcolor{navyblue!10}\textbf{Newly collected}&\\
    \quad\enspace Newly collected Solid Questions & 3,996 \\ 
    \quad\enspace Newly collected Images & 3,792 \\ 
    \midrule
    \rowcolor{navyblue!10}\textbf{Length}&\\
    \quad\enspace Max and Avg question length\,\,\,\,\,\,\,\,\,\,\,\,\,\,\,\,\,\,\,\,\,\,\,\,\,&   1,598\,\,/ 229.8 \\
    \quad\enspace Max and Avg answer length&  501\,\,/\,\,\,\,\,\,\,\,8.9\\
    \quad\enspace Max and Avg image numbers&  8\,\,/\,\,\,\,\,\,\,\,1.1\\
    \bottomrule
  \end{tabular}
}
\captionof{table}{Statistics of Euclid30K, Mathematical Expression, Numeric and Multiple-Choice are the three types of answers.}
\vspace{-1em}
\label{tab:statistics}
\end{table}

%% file: tab/tab_vsibench.tex
\begin{table*}[!t]
\centering
\setlength\tabcolsep{3pt} 
\resizebox{0.961\textwidth}{!}
{
    \begin{tabular}{l|cccc|cccc|c}
        \toprule
        \multirow{2}{*}{\textbf{Methods}}  & \multicolumn{4}{c|}{\textbf{Numerical Question}} & \multicolumn{4}{c|}{\textbf{Multiple-Choice Question}} & \multirow{2}{*}{\textbf{Overall}} \\
        \cmidrule(lr){2-5}\cmidrule(lr){6-9}
         & Obj. Cnt. & Abs. Dist. & Obj. Size & Room Size & Rel. Dist. & Rel. Dir. & Route Plan & Appr. Order &\\
        \midrule
        \rowcolor{navyblue!10}\multicolumn{1}{l|}{\textcolor{black}{\textit{Proprietary Models}}} & & & & & & & & &\\
        GPT-4o~\cite{GPT4o_2024_arXiv}& 46.2 & 5.3 & 43.8 & 38.2 & 37.0 & 41.3 & 31.5 & 28.5 & 34.0 \\
        Gemini-1.5 Pro~\cite{Gemini_1.5_arXiv_2024}& 49.6& 28.8& 58.6 &49.4 &46.0 &48.1& 42.0 &68.0 &48.8\\
        Gemini-2.0 Flash~\cite{google_gemini_2_0_flash_model_card_2025}& 52.4& 30.6 &66.7 &31.8 &56.0 &46.3 &24.5& 55.1&45.4\\
        Gemini-2.5-Pro\cite{Gemini_2.5_arXiv_2025}& -& -& -&-&-&-&-&-&47.8\\
        \midrule
        \rowcolor{navyblue!10}\multicolumn{1}{l|}{\textcolor{black}{\textit{Open-source Models}}} & & & & & & & & &\\
        InternVL2-40B~\cite{Internvl2.5_2025_arXiv}& 34.9 & 26.9 & 46.5 & 31.8 & 42.1 & 32.2 & 34.0 & 39.6 & 36.0 \\
        VILA-1.5-40B~\cite{VILA_CVPR_2023}& 22.4 & 24.8 & 48.7 & 22.7 & 40.5 & 25.7 & 31.5 & 32.9 & 31.2 \\
        LLaVA-OneVision-72B~\cite{LLaVA-OneVision_arXiv-2024}\,\,& 43.5 & 23.9 & 57.6 & 37.5 & 42.5 & 39.9 & 32.5 & 44.6 & 40.2 \\
        LLaVA-Video-72B~\cite{LLaVA-Video_arXiv_2024}& 48.9 & 22.8 & 57.4 & 35.3 & 42.4 & 36.7 & 35.0 & 48.6 & 40.9\\
        \midrule
        \rowcolor{navyblue!10}\multicolumn{1}{l|}{\textcolor{black}{\textit{Spatial Models}}} & & & & & & & & & \\
        M2-Reasoning-7B~\cite{M2reasoning_arXiv_2025}&  41.0& 34.0& 60.9& 55.4& 40.7& 47.3& 29.9& 28.8& 42.3 \\
        Spatial-MLLM-4B~\cite{Spatial-MLLM_2025_arXiv}& 65.3 & 34.8 & 63.1 & 45.1 & 41.3 & 46.2 & 33.5 & 46.3 & 48.4 \\
        \midrule
        \rowcolor{navyblue!10}\multicolumn{1}{l|}{\textcolor{black}{\textit{Qwen2.5VL-series}} } & & & & & & & & & \\
        Qwen2.5VL-3B~\cite{Qwen2.5VL_2025_arXiv}& 35.6 & 23.4 & 34.9 & 16.6 & 34.4 & 40.7 & 26.3 & 21.8 & 29.2 \\
        Qwen2.5VL-Euclid-3B&\UP{38.3}&\UP{26.8}&\UP{35.4}&\UP{22.2}&\UP{37.0}&\UP{43.2}&\UP{36.6}&\DOWN{16.3}&\UP{32.0}\\
        Qwen2.5VL-72B~\cite{Qwen2.5VL_2025_arXiv}& 13.6 & 19.6 & 40.9 & 41.1 & 37.7 & 35.3 & 34.0 & 36.2 & 32.3 \\
        Qwen2.5VL-Euclid-72B  & \UP{22.5} & \UP{27.2} & \UP{55.7} & \UP{43.3} & \UP{44.9} & \UP{37.1} & \DOWN{32.5} & \UP{36.6} & \UP{37.5} \\
        \midrule
        \rowcolor{navyblue!10}\multicolumn{1}{l|}{\textcolor{black}{\textit{Qwen3VL-series}}} & & & & & & & & & \\
        Qwen3VL-4B~\cite{Qwen3VL_2025_arXiv}& 28.5 & 33.0 & 32.6 & 43.5 & 40.3 & 40.0 & 33.0 & 33.2 & 35.5 \\
        Qwen3VL-Euclid-4B & \UP{33.3} & \UP{37.4} & \UP{49.5} & \UP{48.3} & \UP{46.5} & \UP{46.3} & \UP{34.0} & \UP{42.9} & \UP{42.3} \\
        Qwen3VL-8B~\cite{Qwen3VL_2025_arXiv}& 10.8 & 28.1 & 37.9 & 44.1 & 31.3 & 36.4 & 37.1 & 40.3 & 33.3 \\
        Qwen3VL-Euclid-8B & \UP{16.5} & \UP{29.8} & \DOWN{36.0} & \UP{47.7} & \UP{34.5} & \UP{36.6} & \UP{38.1} & \UP{45.1} & \UP{35.5} \\
        Qwen3VL-30B-A3B~\cite{Qwen3VL_2025_arXiv}& 27.4 & 32.3 & 53.6 & 44.0 & 42.1 & 36.2 & 35.5 & 48.9 & 40.0 \\
        Qwen3VL-Euclid-30B-A3B & \UP{33.5} & \UP{37.5} & \UP{64.1} & \DOWN{43.1} & \UP{48.7} & \UP{49.6} & \UP{35.6} & \UP{54.4} & \UP{45.8} \\
        \midrule
        \rowcolor{navyblue!10}\multicolumn{1}{l|}{\textcolor{black}{\textit{RoboBrai2.0-series}}} & & & & & & & & & \\
        RoboBrain2.0-7B~\cite{RoboBrain2.0_2025_TechnicalReport}& 46.0 & 32.7 & 58.9 & 35.9 & 45.9 & 41.5 & 30.9 & 55.2 & 43.0 \\
        RoboBrain2.0-Euclid-7B&\UP{66.4}&\UP{36.9}&\UP{66.3}&\UP{40.5}&\UP{48.3}&\UP{45.3}&\UP{35.6}&\UP{57.8}&\UP{49.6}\\
        RoboBrain2.0-32B~\cite{RoboBrain2.0_2025_TechnicalReport}& 50.5 & 37.0 & 59.2 & 28.4 & 43.2 & 46.1 & 34.5& 39.5 & 43.1 \\
        RoboBrain2.0-Euclid-32B& \UP{59.2} & \UP{39.4} & \UP{63.4} & \UP{47.8} & \UP{48.7} & \UP{47.5} & \DOWN{33.5} & \UP{57.0} & \UP{49.6} \\
        \bottomrule
    \end{tabular}
}
\captionof{table}{\small \textbf{Evaluation on VSI-Bench}~\cite{VSIBench_2025_CVPR}. Qwen2.5VL-Euclid, Qwen3VL-Euclid and RoboBrain2.0-Euclid indicate the Qwen2.5VL~\cite{Qwen2.5VL_2025_arXiv}, Qwen3VL \cite{Qwen3VL_2025_arXiv} and RoboBrain2.0 \cite{RoboBrain2.0_2025_TechnicalReport} trained with GRPO \cite{GRPO_DeepSeekMath_arXiv_2024} on the Euclid30K dataset.}
\label{tab:vsibench}
\end{table*}

%% file: sec/4_expeiment.tex
\section{Experiment}

In this section, we evaluate the zero-shot generalisation of models trained on geometry data to spatial intelligence tasks, using four benchmarks: Super CLEVR, Omni3D Bench, VSI Bench, and MindCube.
We also conduct a causal ablation study that contrasts models trained on equal amounts of geometry data and spatial data.
Additional experiments, along with detailed settings, prompt templates, and dataset settings, are provided in the \appref{sec:appendix_exp}.
Moreover, more visualization and analysis of results are presented in \appref{sec:appendix_vis} and Figs.~\ref{fig:app1}–-\ref{fig:comp8} in the Appendix.
\subsection{Comparisons on VSI-Bench}\label{ssec:exp_vsi}

As noted in VSIBench, spatial reasoning ability is the primary bottleneck limiting MLLM performance on the VSI-Bench test~\cite{VSIBench_2025_CVPR}. 
Therefore, to better demonstrate how models perceive scenes and perform spatial reasoning, and to verify whether they genuinely acquire spatial intelligence from geometric knowledge, we deviate from the original VSI-Bench setup, which uses prompts such as "\textit{Please answer the question using a single word or phrase}" and constrains the maximum response length to 16 tokens.
Instead, we follow the prompt configuration described in RoboBrain2.0~\cite{RoboBrain2.0_2025_TechnicalReport} Sec.~B, which encourages the model to first reason about the problem before providing an answer, and we set the maximum response length to 1024 tokens.
For fairness, all compared models (baselines and Euclid30K‑fine‑tuned variants) use the same prompt template and a 1024‑token response budget.
This setup allows us to observe the model's intermediate reasoning process and assess whether it has internalized transferable spatial priors from Euclid30K training.

\cref{tab:vsibench} shows that Euclid30K fine-tuning yields an average accuracy improvement of 5.2\% across all evaluated model variants in the Qwen2.5VL, Qwen3VL, and RoboBrain2.0 families.
The effect is most pronounced on RoboBrain2.0, where the Euclid-trained 7B and 32B versions reach 49.6\% overall, outstripping the best open-source reference (Spatial-MLLM-4B at 48.4\%) and also surpassing the strongest proprietary baselines reported by VSI-Bench (Gemini-1.5 Pro and Gemini-2.5-Flash-preview-04-17, 47.8\% and 48.8\%).
Furthermore, the visualization examples in the \appref{ssec:app_res} demonstrate that these models have internalized fundamental spatial knowledge, such as perspective scaling, size estimation, and relational reasoning, by engaging with geometric knowledge.

\subsection{Comparisons on SuperClevr \& Omni3DBench}\label{ssec:exp_base}

\input{tab/val_and_test}

\cref{tab:val_and_test} demonstrates that the fine-tuned Euclid30K model achieves improved accuracy on both SuperClevr and Omni3DBench, two classic spatial intelligence datasets.
For Qwen2.5VL, the 3B model rises from 70.0 to 75.2 on Super-CLEVR and from 24.7 to 26.5 on Omni3D-Bench; the 72B model adds 10.5 percentage points on Super-CLEVR and about 2.5 on Omni3D-Bench. 
Qwen3VL shows a similar pattern.
The dense 4B variant improves by 5.8 and 4.0 points on the two datasets, and the MoE 30B-A3B improves by 6.1 and 2.2 points.
Overall, Euclid30K delivers substantial accuracy gains, demonstrating that the geometry-centric curriculum provides broadly transferable spatial priors rather than mere data volume benefits.

\input{tab/tab_mindcube}

The substantial performance gains observed for RoboBrain2.0 after Euclid30K fine-tuning are plausibly linked to its prior exposure: its SFT corpus includes substantial spatial-intelligence data, which likely instills implicit, intuitive notions about spatial phenomena.
Training on Euclid30K then supplies explicit first-principles geometric reasoning, formalizing and consolidating those earlier intuitions, thereby enabling the model to generalize more rapidly to downstream tasks.
This synergy between implicit spatial exposure and explicit geometric grounding may account for the pronounced performance improvements.

\subsection{Comparisons on MindCube}\label{ssec:exp_mindcube}

\input{tab/ab_vsi}

As shown in \cref{tab:mindcube}, fine-tuning on Euclid30K improves the overall MindCube accuracy across all variants in the three model families, with an average gain of 6.7 percentage points. 
It is also worth noting that models trained using Euclidean geometry datasets outperform existing spatial models (most of which are also based on Qwen backbones, but trained on a larger spatial corpus) in terms of generalization ability. 
For example, Spatial-MLLM combines the Qwen 2.5-VL-3B and VGGT~\cite{vggt_2025_CVPR} backbones, trains on 120K spatial samples, and scores 32.1\% on MindCube.
In contrast, Qwen2.5VL-Euclid-3B, trained with only 30K geometric datas, scored 38.9\% on MindCube, representing a relative improvement of 6.8 percentage points. 
These results suggest that learning accurate Euclidean priors from a compact geometry course provides more transferable spatial knowledge than extending generalized spatial data alone.

\subsection{Ablation Study}\label{ssec:ablation}

To isolate the contribution of our Euclid30K dataset from the potential reasoning enhancements provided by the GRPO algorithm, we conducted a causal ablation study. 
Specifically, we randomly sampled a subset equal in size to Euclid30K on the non-geometric spatial intelligence dataset Clevr-CoGenT~\cite{Clevr_2017_CVPR} and used the exact same GRPO setup to train Qwen2.5VL, Qwen3VL and RoboBrain2.0.
This design ensures that performance gains after training on geometric data can be directly attributed to the fact that the geometric task as surrogate task for spatial intelligence, rather than due to the effects of GRPO or data incrementation.

\cref{ab:vsibench} shows that models trained on Euclid30K achieve markedly higher overall accuracy than those fine-tuned on the equal-sized Clevr-CoGenT split.
Specifically, Euclid30K training improves average accuracy from 38.5\% to 44.3\% (+5.8 points), whereas Clevr-CoGenT training yields a smaller improvement from 38.5\% to 41.4\% (+2.9 points).
This demonstrates that the performance gains from Euclid30K exceed those attributable to additional data or GRPO-induced generalization alone, confirming that structured geometric knowledge provides a more transferable foundation for spatial reasoning.
Ablation results for additional model sizes are provided in \appref{ssec:app_ablation}.

%% file: tab/val_and_test.tex
\begin{table}[!t]
\centering
\renewcommand{\arraystretch}{1.2}
\setlength\tabcolsep{3pt} 
\resizebox{0.85\linewidth}{!}
{
  \begin{tabular}{l|cc}
    \toprule
    \textbf{Methods}   & \textbf{SuperClevr} &  \textbf{Omni3DBench}  \\
    \midrule
    \rowcolor{navyblue!10}\textit{Qwen2.5VL-series}&& \\
    Qwen2.5VL-3B  & 70.0 & 24.7  \\
    Qwen2.5VL-Euclid-3B   &\UP{75.2}&\UP{26.5}  \\
    Qwen2.5VL-72B & 72.6 & 30.4  \\
    Qwen2.5VL-Euclid-72B    &\UP{83.1}&\UP{32.9}\\
    \midrule
    \rowcolor{navyblue!10}\textit{Qwen3VL-series}&& \\
    Qwen3VL-4B  & 55.4 & 27.7  \\
    Qwen3VL-Euclid-4B   &\UP{61.2}&\UP{31.7}  \\
    Qwen3VL-8B  & 48.3 & 34.0  \\
    Qwen3VL-Euclid-8B    &\UP{49.0}&\UP{35.0}\\
    Qwen3VL-30B-A3B  & 64.1 & 36.7  \\
    Qwen3VL-Euclid-30B-A3B\,\,\,\,   &\UP{70.2}&\UP{38.9}  \\
    \midrule
    \rowcolor{navyblue!10}\textit{RoboBrain2.0-series}&& \\
    RoboBrain2.0-7B    & 47.4 & 14.2  \\
    RoboBrain2.0-Euclid-7B   & \UP{85.2} & \UP{21.2}  \\
    RoboBrain2.0-32B     & 59.5 & 34.8\\
    RoboBrain2.0-Euclid-32B   & \UP{75.6} & \UP{36.8} \\
    \bottomrule
  \end{tabular}
}
\captionof{table}{\small \textbf{Evaluation on SuperClevr}~\cite{Super-CLEVR_CVPR_2023} \&\textbf{ Omni3D Bench}~\cite{Omni3DBench_arXiv_2025}.  }\label{tab:val_and_test}
\vspace{-3mm}
\end{table}

%% file: tab/tab_mindcube.tex
\begin{table}[!t]
  \centering
  \renewcommand{\arraystretch}{1.2}
  \setlength\tabcolsep{3pt}
  \resizebox{0.95\linewidth}{!}
  {
    \begin{tabular}{l|ccc|c}
      \toprule
      \textbf{Methods}                                          & \textbf{Rotation} & \textbf{Among}  & \textbf{Around} & \textbf{Overall} \\
      \midrule
      \rowcolor{navyblue!10}\textit{Proprietary Models}         & \rule{0pt}{1em}   & \rule{0pt}{1em} & \rule{0pt}{1em} & \rule{0pt}{1em}  \\
      GPT-4o~\cite{GPT4o_2024_arXiv}                            & 32.7              & 40.2            & 29.2            & 38.8             \\
      Claude-4-Sonnet~\cite{Claude-4-Sonnet-20250514}           & 48.4              & 44.2            & 47.6            & 44.8             \\
      \midrule
      \rowcolor{navyblue!10}\textit{Spatial Models}             &                   &                 &                 &                  \\
      RoboBrain1.0-7B~\cite{Robobrain_arXiv_2025}               & 35.8              & 38.3            & 29.5            & 37.4             \\
      SpaceMantis~\cite{SpatialVLM_arXiv_2024}                  & 37.7              & 21.3            & 29.3            & 22.8             \\
      Space-Qwen~\cite{SpatialVLM_arXiv_2024}                   & 38.0              & 33.7            & 26.3            & 33.3             \\
      Spatial-MLLM~\cite{Spatial-MLLM_2025_arXiv}               & 38.4              & 20.9            & 32.8            & 32.1             \\
      \midrule
      \rowcolor{navyblue!10}\textit{Qwen2.5VL-series}           &                   &                 &                 &                  \\
      Qwen2.5VL-3B~\cite{Qwen2.5VL_2025_arXiv}                  & 14.3              & 22.8            & 24.1            & 20.4             \\
      Qwen2.5VL-Euclid-3B                                       & \UP{33.5}         & \UP{43.0}       & \UP{40.0}       & \UP{38.9}        \\
      Qwen2.5VL-72B~\cite{Qwen2.5VL_2025_arXiv}                 & 31.5              & 31.4            & 30.6            & 31.2             \\
      Qwen2.5VL-Euclid-72B                                      & \UP{43.0}         & \UP{35.6}       & \UP{31.6}       & \UP{36.7}        \\
      \midrule \rowcolor{navyblue!10}\textit{Qwen3VL-series}    &                   &                 &                 &                  \\
      Qwen3VL-4B~\cite{Qwen3VL_2025_arXiv}                      & 19.3              & 25.3            & 33.7            & 26.1             \\
      Qwen3VL-Euclid-4B                                         & \UP{31.7}         & \UP{29.2}       & \UP{38.0}       & \UP{33.0}        \\
      Qwen3VL-8B~\cite{Qwen3VL_2025_arXiv}                      & 25.3              & 32.4            & 44.7            & 34.2             \\
      Qwen3VL-Euclid-8B                                         & \UP{40.0}         & \UP{35.1}       & \UP{48.0}       & \UP{41.0}        \\
      Qwen3VL-30B-A3B~\cite{Qwen3VL_2025_arXiv}                 & 37.7              & 32.3            & 49.3            & 39.8             \\
      Qwen3VL-Euclid-30B-A3B                                    & \UP{39.6}         & \UP{33.8}       & 48.6            & \UP{40.7}        \\ 
      \midrule
      \rowcolor{navyblue!10}\textit{RoboBrain2.0-series}       &                   &                 &                 &                  \\
      RoboBrain2.0-7B~\cite{RoboBrain2.0_2025_TechnicalReport}  & 39.4              & 38.8            & 38.6            & 38.9             \\
      RoboBrain2.0-Euclid-7B                                    & \DOWN{36.0}       & \UP{46.5}       & \DOWN{36.2}     & \UP{39.4}        \\
      RoboBrain2.0-32B~\cite{RoboBrain2.0_2025_TechnicalReport} & 21.2              & 35.4            & 31.0            & 29.2             \\
      RoboBrain2.0-Euclid-32B                                   & \UP{38.9}         & \UP{37.8}       & \UP{33.8}       & \UP{36.8}        \\
      \bottomrule
    \end{tabular}
  }
  \captionof{table}{\small \textbf{Evaluation on MindCube}~\cite{MindCube_arXiv_2025}. }\label{tab:mindcube}
  \vspace{-3mm}
\end{table}

%% file: tab/ab_vsi.tex
\begin{table*}[!t]
\centering
\setlength\tabcolsep{3pt} 
\resizebox{0.95\textwidth}{!}
{
    \begin{tabular}{l|cccc|cccc|c}
        \toprule
        \multirow{2}{*}{\textbf{Methods}}  & \multicolumn{4}{c|}{\textbf{Numerical Question}} & \multicolumn{4}{c|}{\textbf{Multiple-Choice Question}} & \multirow{2}{*}{\textbf{Overall}} \\
        \cmidrule(lr){2-5}\cmidrule(lr){6-9}
         & Obj. Cnt. & Abs. Dist. & Obj. Size & Room Size & Rel. Dist. & Rel. Dir. & Route Plan & Appr. Order &\\
        \midrule
        Qwen2.5VL-72B  & 13.6 & 19.6 & 40.9 & 41.1 & 37.7 & 35.3 & \textbf{34.0} & 36.2 & 32.3 \\
        Qwen2.5VL-Space-72B & 15.6 & 24.8 & 40.7 & 41.4 & 43.4 & \textbf{37.8}& 29.4 & 33.5 & 33.2\\
        Qwen2.5VL-Euclid-72B  &  \textbf{22.5} & \textbf{27.2} & \textbf{55.7} &  \textbf{43.3} & \textbf{44.9} &  37.1 &  32.5 &  \textbf{36.6} & \textbf{37.5}\\
        \midrule
        Qwen3VL-30B-A3B & 27.4 & 32.3 & 53.6 & \textbf{44.0} & 42.1 & 36.2 & 35.5 & 48.9 & 40.0 \\
        Qwen3VL-Space-30B-A3B & 29.9 & 36.8 & 58.0 & 43.4 & 47.9 & 49.1 & \textbf{35.9} & 52.7 & 44.2\\
        Qwen3VL-Euclid-30B-A3B & \textbf{33.5} & \textbf{37.5} & \textbf{64.1} & 43.1 & \textbf{48.7} & \textbf{49.6} & 35.6 & \textbf{54.4} & \textbf{45.8} \\
        \midrule
        RoboBrain2.0-32B& 50.5 & 37.0 & 59.2 & 28.4 & 43.2 & 46.1 & \textbf{34.5}& 39.5 & 43.1 \\
        RoboBrain2.0-Space-32B& 58.0 & 36.9 & 62.2 &\textbf{47.8} & 46.9 & 44.5& 34.0& 42.1 & 46.7\\
        RoboBrain2.0-Euclid-32B& \textbf{59.2} &  \textbf{39.4} & \textbf{63.4} & \textbf{47.8}& \textbf{48.7} & \textbf{47.5} & 33.5 & \textbf{57.0} & \textbf{49.6} \\
        \bottomrule
    \end{tabular}
}
\captionof{table}{\small \textbf{Ablation experiment on VSI-Bench}~\cite{VSIBench_2025_CVPR}. We compare training a model on a 30K subset of the spatial intelligence dataset Clevr-CoGenT v.s. the geometric dataset Euclid30K to verify that the geometric dataset serves as a surrogate task to improve the spatial intelligence capabilities of the model. \textbf{Bolding} indicates the best score within each model type.}
\vspace{-0.5em}
\label{ab:vsibench}
\end{table*}

%% file: sec/5_discussion.tex
\section{Discussion}

\noindent{\textbf{Supervised Fine-tuning (SFT)}}. For mathematical problems, collecting step-by-step solution annotations is substantially more expensive than answer-only labels.
Accordingly, Euclid30K does not currently include process annotations, which makes it less suited to SFT routines that rely on explicit reasoning traces.
Nevertheless, geometry QA can still serve as a surrogate task for spatial intelligence under SFT.\@
In Appendix \appref{ssec:sft}, we report SFT results using other geometry datasets that provide process supervision. 
As future work, we plan to extend Euclid30K with solution-process annotations to better support SFT.\@

\noindent{\textbf{Model-Specific Performance Variations}}.
Although geometry tasks provide beneficial Euclidean priors that enhance spatial reasoning across \cref{tab:vsibench,tab:val_and_test,tab:mindcube}, improvements exhibit model-specific variation. 
We offer several insights into this phenomenon. 
Performance gains appear linked to characteristics inherited from the model's earlier training stages. 
Specifically, when a model already possesses rudimentary spatial concepts, training on Euclid30K refines and consolidates that knowledge through first-principles geometric reasoning, yielding larger improvements (\textit{e.g.}, RoboBrain2.0).
Conversely, if a model must discover these first principles from scratch using only geometric data, gains tend to be more gradual.
We provide additional experiments exploring this phenomenon in \appref{ssec:var}.


%% file: sec/6_conclusion.tex
\section{Conclusion}

This study shows that using geometry as a surrogate task provides an alternative way to achieve transferable Spatial Perception and Reasoning. 
MLLMs trained on planar and solid geometry learn basic Euclidean priors that can be transferred to multiple spatial tasks without extra fine-tuning. 
Training solely on our Euclid30K yields consistent, significant gains across four unseen benchmarks for seven model. 
These generalization gains validate our core hypothesis that learning basic Euclidean geometric principles is an effective strategy for developing transferable spatial skills.

%% file: sec/X_suppl.tex
\clearpage
\appendix
\maketitlesupplementary{}

\begingroup
  \setcounter{tocdepth}{2} 
  \renewcommand{\contentsname}{Supplementary Contents}
  \phantomsection
  \pdfbookmark[1]{Supplementary Contents}{supp_toc}
  \hypersetup{linkcolor=black}
  \tableofcontents
\endgroup

\section{Proof of Standard Domain-Adaptation Bound}
\label{sec:appendex_rationale}

This section provides the complete proof for the standard domain-adaptation bound introduced in \cref{ssec:motivation}.
We follow the same notation as in the main text. 

Let \(h^*\in\mathcal{H}\) be the (ideal) hypothesis defined by
\begin{equation}
  h^* = \arg\min_{h\in\mathcal{H}}\big(\epsilon_S(h) + \epsilon_T(h)\big),
  \label{eq:appendix_define_h-star}
\end{equation}
and denote \(\epsilon_{\text{ideal}} = \epsilon_S(h^*) + \epsilon_T(h^*)\).
By the definition of \(d_{\mathcal{H}\Delta\mathcal{H}}\), we have
\begin{equation}
 \big|\epsilon_S(h,h^*)-\epsilon_T(h,h^*)\big| \le \tfrac{1}{2} d_{\mathcal{H}\Delta\mathcal{H}}(\mathcal{D}_S,\mathcal{D}_T).
 \label{eq:appendix_discrepancy-lemma}
\end{equation}

Then, by the triangle inequality and \cref{eq:appendix_discrepancy-lemma}, we have:
\begin{equation}
\begin{aligned}
 \epsilon_T(h)
 &\le \epsilon_T(h^*) + \epsilon_T(h,h^*) \\
 &= \epsilon_T(h^*) + \epsilon_S(h,h^*) + \big(\epsilon_T(h,h^*)-\epsilon_S(h,h^*)\big) \\
 &\le \epsilon_T(h^*) + \epsilon_S(h,h^*) + \big| \epsilon_T(h,h^*)-\epsilon_S(h,h^*) \big| \\
 &\le \epsilon_T(h^*) + [\epsilon_S(h) + \epsilon_S(h^*)] + \tfrac{1}{2} d_{\mathcal{H}\Delta\mathcal{H}}(\mathcal{D}_S,\mathcal{D}_T) \\
 &= \epsilon_S(h) + \epsilon_{ideal} + \tfrac{1}{2}\, d_{\mathcal{H}\Delta\mathcal{H}}(\mathcal{D}_S,\mathcal{D}_T),
\end{aligned}
\label{eq:appendix_supp-da-bound}
\end{equation}
The above proof refers to~\cite{BenDavid2010} and~\cite{mansour2009domain}.

In the setting considered in this paper we examine generalisation from (i) formal geometry to broad spatial intelligence and (ii) spatial sub-skills to broader spatial intelligence.
Given continued scaling of model capacity and data, along with advances in training methodologies, it is reasonable to anticipate the emergence of a sufficiently strong hypothesis $h^*$ for which the source-target joint errors in both regimes satisfy:
\begin{equation}
\epsilon_S(h^*) \approx \epsilon_T(h^*) \approx 0,
\end{equation}
\textit{i.e.}, the error magnitudes on geometry tasks and spatial-intelligence tasks (and likewise on spatial sub-tasks and full spatial intelligence) become simultaneously negligible.
When this asymptotic condition holds for both sides of our comparisons, the corresponding ideal terms $\epsilon_{\text{ideal}}$ in each bound are effectively $0$, making their omission justified.
Consequently, Eq.~\eqref{eq:appendix_supp-da-bound} simplifies to:
\begin{equation}
\epsilon_T(h) \lesssim \epsilon_S(h) + \tfrac{1}{2} d_{\mathcal{H}\Delta\mathcal{H}}(\mathcal{D}_S,\mathcal{D}_T),
\end{equation}
Therefore, if $d_{\mathcal{H}\Delta\mathcal{H}}(\mathcal{D}_S,\mathcal{D}_T)$ is sufficiently small, the population gap $\epsilon_T(h) - \epsilon_S(h)$ also becomes small, which allows the source distribution to serve as a reliable surrogate for the target distribution \cite{liu2025spatial, pmlr-v139-acuna21a,liu2025mirepnet}.

\section{Evidence from Educational Psychology}
\label{sec:edu-psych}
Complementing the domain-adaptation view in the previous subsection, we now present evidence from educational psychology that echoes the cognitive-science perspective on the generality of geometric knowledge in perception and reasoning~\cite{lake2017building,feldman2003visual}.

There is extensive evidence in educational psychology that geometry problem solving is closely related to spatial intelligence, can serve as an informative indicator of spatial ability, and can be used to improve it through targeted practice.

First, numerous correlational studies document a substantive link between geometric and spatial reasoning.
Kyaw and Vid\'akovich report a moderate positive correlation between teachers' geometric and spatial reasoning (\(r=0.47\)), with 3D matching and measurement tasks predicting spatial scores~\cite{KyawVidakovich2025}.
In STEM and graphical education, higher spatial ability is associated with better problem-solving performance and more effective strategies~\cite{buckley2019investigating}.
Newcombe and Frick emphasize that spatial representations and transformations are central cognitive resources that support reasoning in domains that are not obviously spatial—for example, through the use of graphs and diagrams~\cite{newcombe2010early}.

Second, several studies show that performance on geometry tasks is a sensitive proxy for spatial ability.
Analyses of middle-school students reveal that geometry skills and error patterns systematically vary with spatial-intelligence levels~\cite{riastuti2017analysis,riastuti2017students}.
Differences in dominance between logical-mathematical and visual-spatial intelligence yield distinct pathways for geometric reasoning, further tying geometry problem solving to spatial constructs~\cite{aziz2020students}.
These results support the use of geometry assessments as indicators of students' spatial proficiency.

Third, intervention studies demonstrate that providing structured geometric activities can improve spatial intelligence.
Programmatic practice with polyhedra and computer-generated spatial problems yields measurable gains~\cite{aszalos2004can}.
Geometrical-mechanical intelligence games, implemented in quasi-experiments with pre/post testing, significantly enhance spatial visualisation and spatial relations skills~\cite{SutcuOral2020}.
Overall, the balance of evidence indicates that well-designed geometric practice is an effective means to cultivate spatial abilities.

Moreover, neuroscience research reveals that early exposure to Euclidean geometric structures fundamentally shapes spatial representations.
Studies examining hippocampal activity in rodents reared in spherical versus cuboid environments demonstrate that experience with canonical Euclidean features (edges, corners, planes) enriches the repertoire of preconfigured neuronal patterns and enhances the brain's ability to discriminate between distinct spatial layouts~\cite{Euclidean_experience_2024_NatureCommunications}.
While these findings originate from animal models, they provide convergent biological evidence that geometric experience during development can refine spatial coding mechanisms—a principle that may extend to learning systems more broadly.

Taken together, these findings motivate our surrogate-task choice.
Our results suggest that the same relationship generalises beyond human learners to large multimodal models: training on formal geometry induces domain-invariant structure that transfers to diverse spatial-intelligence benchmarks.
This observation is consistent with the domain-adaptation analysis in the previous subsection and provides an educational-psychology rationale for our geometry-first curriculum.

\section{Detailed Experimental Setup}
\label{sec:appendix_exp}

This section summarises the key hyperparameters, evaluation settings, prompt templates, and datasets settings used throughout the paper.  

\subsection{Training setup}
In this paper, we follow the default settings of VeRL~\cite{verl_hybridflow_ECCS_2025} and EasyR1~\cite{easyr1_2025_github} to train the Qwen2.5-VL series, Qwen3-VL series, and the RoboBrain2.0 series. 
Specifically, we train for 10 epochs in 64 NVIDIA H100 GPUs using Adam optimizer with a learning rate of $1\times10^{-6}$ and a weight decay of $1\times10^{-2}$.
In GRPO, we perform 8 rollouts per question and set the default sampling temperature to 1. The KL divergence coefficient $\beta$ in Eq.~\ref{eq:loss2} is set to $1\times10^{-2}$. 

Unless stated otherwise, we fix the random seed at \texttt{1} to guarantee determinism.  
We adopt a context window of 1024 tokens for both the prompt and the response, and use a rollout batch of 512 samples.  
The actor network updates with a global batch size of 128 and a maximum gradient norm of 1.0.  
Images are resized so that the total pixel count lies between $512 \times512$ and $2048 \times2048$.  
All remaining hyper-parameters, including PPO clip ratio, learning-rate schedule, and parallelism settings, follow the default EasyR1 recipe and can be found in our GitHub repository.

\subsection{Test setup}
Inference is conducted with the lmms-eval toolkit~\cite{LMMs-Eval_arXiv_2024} to ensure consistent decoding across models.  
In the test, to ensure the reproducibility of the results, we follow VSIBench~\cite{VSIBench_2025_CVPR} and MindCube~\cite{MindCube_arXiv_2025} to set the temperature to 0.
Finally, to ensure that the model performs sufficient spatial inference, we set the maximum generation length of model responses at 1024 tokens.
For reproducibility, detailed testing scripts are provided in our GitHub repository.

\subsection{Prompt templates}
\vspace{0.3em}
\noindent\textbf{Euclid-tuned models}.\quad  
During both training and evaluation, we use the following template:

\hypbox{Euclid-tuned Models Prompt Template}{%
You FIRST think about the reasoning process as an internal monologue and then provide the final answer. The reasoning process MUST BE enclosed within \textless think\textgreater\,\textless /think\textgreater~tags. The final answer MUST BE put in \textbackslash boxed\{\}.
}

\noindent\textbf{Baseline variants}.\quad  
RoboBrain2.0 expects the answer inside \textless~answer\textgreater\,\textless~/answer\textgreater~tags; we therefore replace the last line with, like:  
\hypbox{Vanilla RoboBrain2.0 Prompt Template}{%
You FIRST think about the reasoning process as an internal monologue and then provide the final answer. The reasoning process MUST BE enclosed within \textless think\textgreater\,\textless /think\textgreater~tags. The final answer MUST BE put in \textless answer\textgreater\,\textless /answer\textgreater.
}

Because Qwen2.5VL-Instruct and Qwen3VL-Instruct was tuned with supervised instruction data that often begins with phrases like “think step by step,” keeping the same cue in your evaluation prompt aligns the test-time input with the style encountered during training. This consistency helps the model interpret the prompt as intended and reduces the risk of unexpected formatting effects.
\hypbox{Vanilla QwenVL-Instruct Prompt Template}{%
You FIRST think step by step and then provide the final answer. The final answer MUST BE put in \textbackslash boxed\{\}.
}

During evaluation, to mitigate prompt-template bias, we run each model with both its native instruction-style prompt and the unified Euclid reasoning template, and report the better-performing variant.
When processing VSI-Bench tasks, we make minor adjustments to the above templates to follow the benchmark's original settings as closely as possible and ensure consistent results—for example, we prepend “These are frames of a video.” to every prompt.

\begin{figure*}[ht]
\centering
\includegraphics[width=1\linewidth]{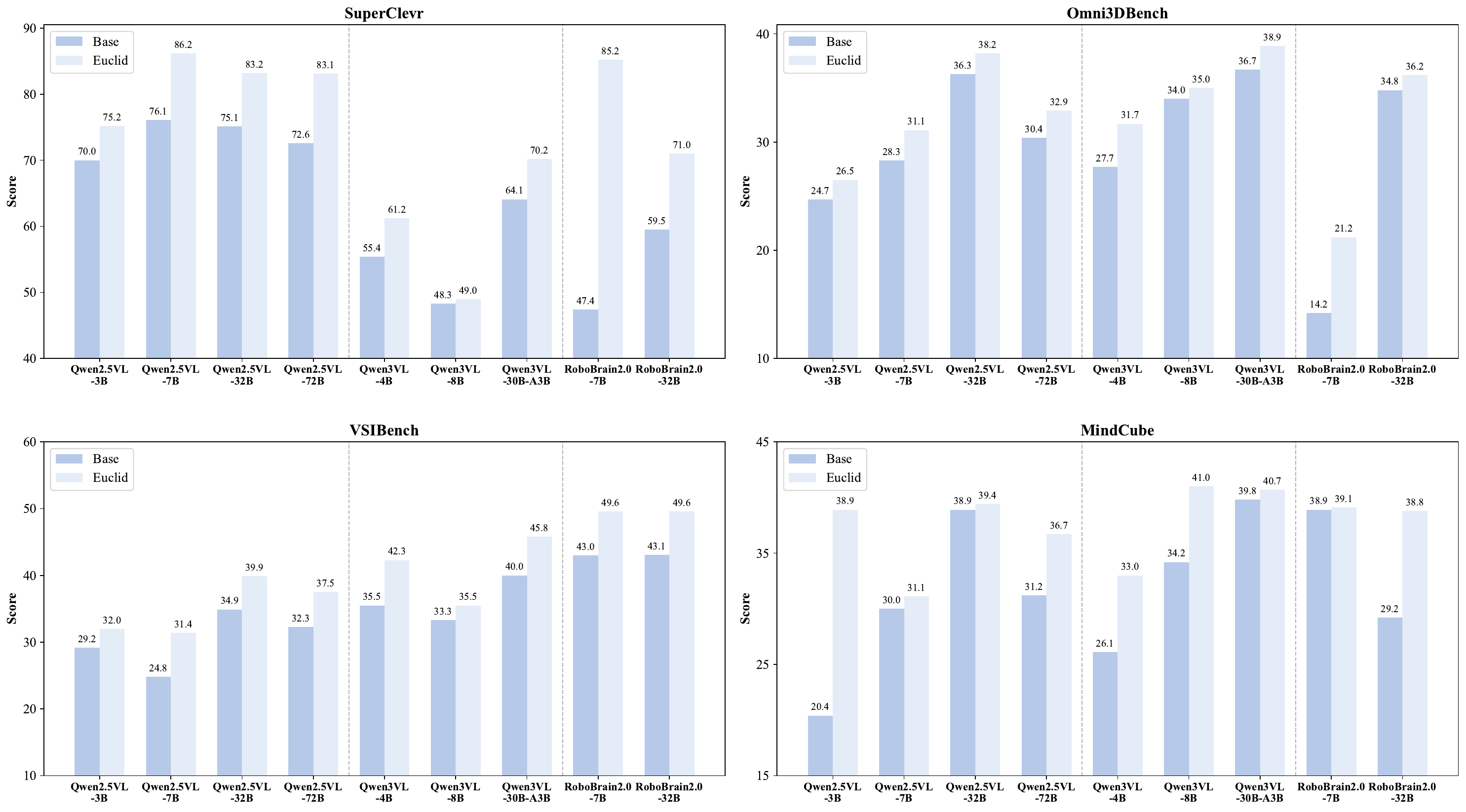}%
\caption{\textbf{Performance improvement} on SuperClevr~\cite{Super-CLEVR_CVPR_2023}, Omni3DBench~\cite{Omni3DBench_arXiv_2025}, VSIBench~\cite{VSIBench_2025_CVPR}, and MindCube~\cite{MindCube_arXiv_2025} after the model has been trained on Eculid30K.}
\label{fig:app1}
\end{figure*}

\subsection{Dataset Setup}\label{sec:appendix_dataset_setup}
In this subsection, we provide an introduction and configuration details for the dataset used in the main page.

\noindent\textbf{Setup in VSI-Bench}.
VSI-Bench~\cite{VSIBench_2025_CVPR} contains more than 5,130 egocentric videos question-answer pairs sourced from ARKitScenes\cite{ARKitScene_2021_NIPS}, ScanNet~\cite{ScanNet_2017_CVPR}, and ScanNet++ \cite{ScanNet++_2023_ICCV}. The task types are divided into numerical question tasks (e.g., object counting, absolute distance estimation, object size estimation, and room size estimation) and multiple choice tasks (e.g., relative distance estimation, relative direction reasoning,  route planning, and spatiotemporal appearance‑order). 
For the evaluation metrics, we align with the VSIBench setting. 
In addition, for the Qwen2.5VL-series and RoboBrain2.0-series, we use 32 frames uniformly sampled from the scene video as input frames in the inference process. 

\noindent\textbf{Setup in Super‑CLEVR and Omni3D‑Bench}. 
Super‑CLEVR~\cite{Super-CLEVR_CVPR_2023} contains a 5,000-image test split that probes how well a model handles changes in visual complexity, concept distribution, and composition, making it a strong measure of two‑dimensional spatial reasoning. Omni3D‑Bench~\cite{Omni3DBench_arXiv_2025} adds 500 questions to the Omni3D dataset, each requiring a model to locate objects in three‑dimensional space and estimate their relative distances and sizes. Together, these benchmarks test both planar and volumetric aspects of spatial understanding, providing complementary evidence of a model’s geometric competence.
For the evaluation metrics, we follow the settings of VSIBench~\cite{VSIBench_2025_CVPR}. Specifically, we calculate mean relative accuracy (MRA) across confidence thresholds  $\mathcal{C} = \{0.5,0.55 \dots, 0.95\}$ for the numerical question tasks and report exact‑match accuracy for multiple‑choice tasks. 

\noindent\textbf{Setup in MindCube}. 
MindCube~\cite{MindCube_arXiv_2025} is a recent benchmark crafted to scrutinize the spatial-reasoning capabilities of VLMs under partial observability and dynamic viewpoints, challenging the VLM to maintain object consistency across viewpoints and to reason about occluded or invisible elements. MindCube defines three canonical camera trajectories: Rotation (camera stays in place but rotates to look around; 1,081 samples), Around (camera moves around objects in a circular path; 1,869 samples), and Among (camera moves among objects in a circular path; 18,204 samples). Since all questions follow a multiple‑choice format, we evaluate models by exact‐match accuracy between the predicted option and the ground‑truth answer.

\section{More Experiment and Visualization}\label{sec:appendix_vis}

\input{tab/appendix_abvsi.tex}

\subsection{More experiments about main results}\label{ssec:app_res}

To present the quantitative gains more intuitively, Fig.~\ref{fig:app1} plots the base models and their Euclid30K-tuned counterparts side by side.
The light bars show consistent accuracy improvements on Super-CLEVR~\cite{Super-CLEVR_CVPR_2023}, Omni3D-Bench~\cite{Omni3DBench_arXiv_2025}, VSI-Bench~\cite{VSIBench_2025_CVPR}, and MindCube~\cite{MindCube_arXiv_2025}, confirming that a compact geometry curriculum injects transferable spatial priors across both Qwen2.5VL, Qwen3VL and RoboBrain2.0 families.
Additionally, we include results for Qwen2.5VL-7B and Qwen2.5VL-32B in the Fig.~\ref{fig:app1}, which exhibit consistent improvements across all four benchmarks following the same geometric surrogate-task training, reaffirming that geometry serves as an effective surrogate task for spatial intelligence and further validating the robustness and generality of this approach.

\begin{figure*}[h]
    \centering
    \includegraphics[width=1\linewidth]{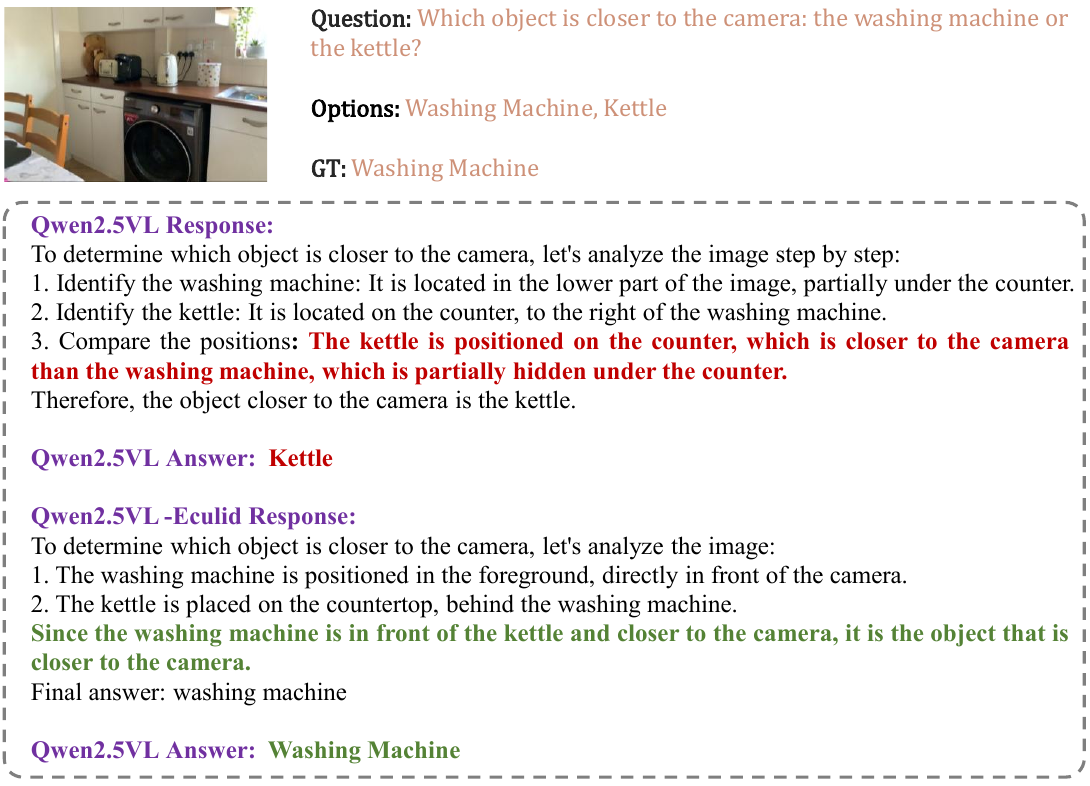}%
    \caption{The response and final answer for Qwen2.5VL-7B~\cite{Qwen2.5VL_2025_arXiv} and Qwen2.5VL-Eculid-7B in Omni3DBech~\cite{Omni3DBench_arXiv_2025}.}
    \label{fig:comp1}
\end{figure*}

\begin{figure*}[h]
    \centering
    \includegraphics[width=1\linewidth]{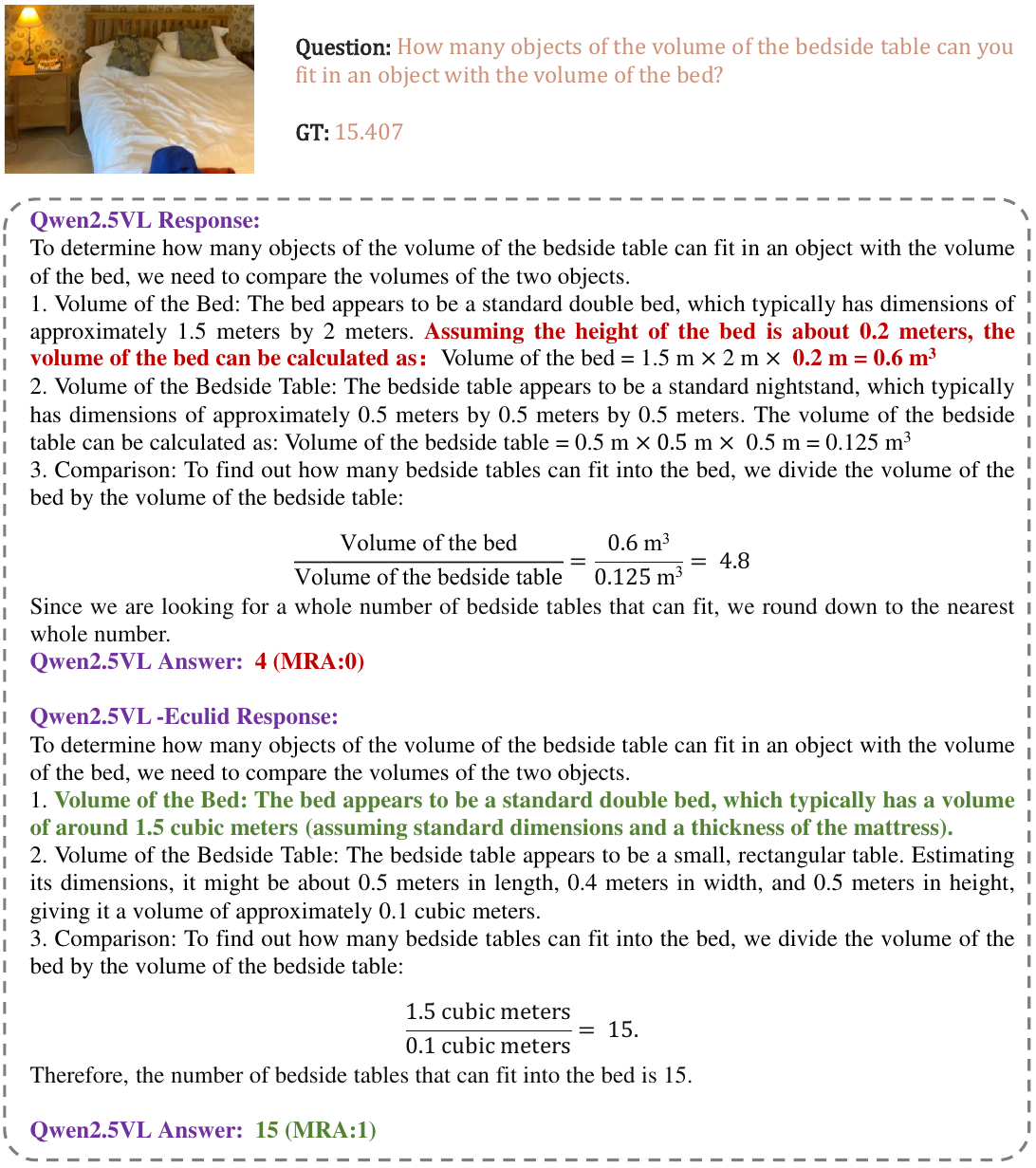}%
    \caption{The response and final answer for Qwen2.5VL-7B~\cite{Qwen2.5VL_2025_arXiv} and Qwen2.5VL-Eculid-7B in Omni3DBech~\cite{Omni3DBench_arXiv_2025}.}
    \label{fig:comp2}
\end{figure*}

\begin{figure*}[h]
    \centering
    \includegraphics[width=1\linewidth]{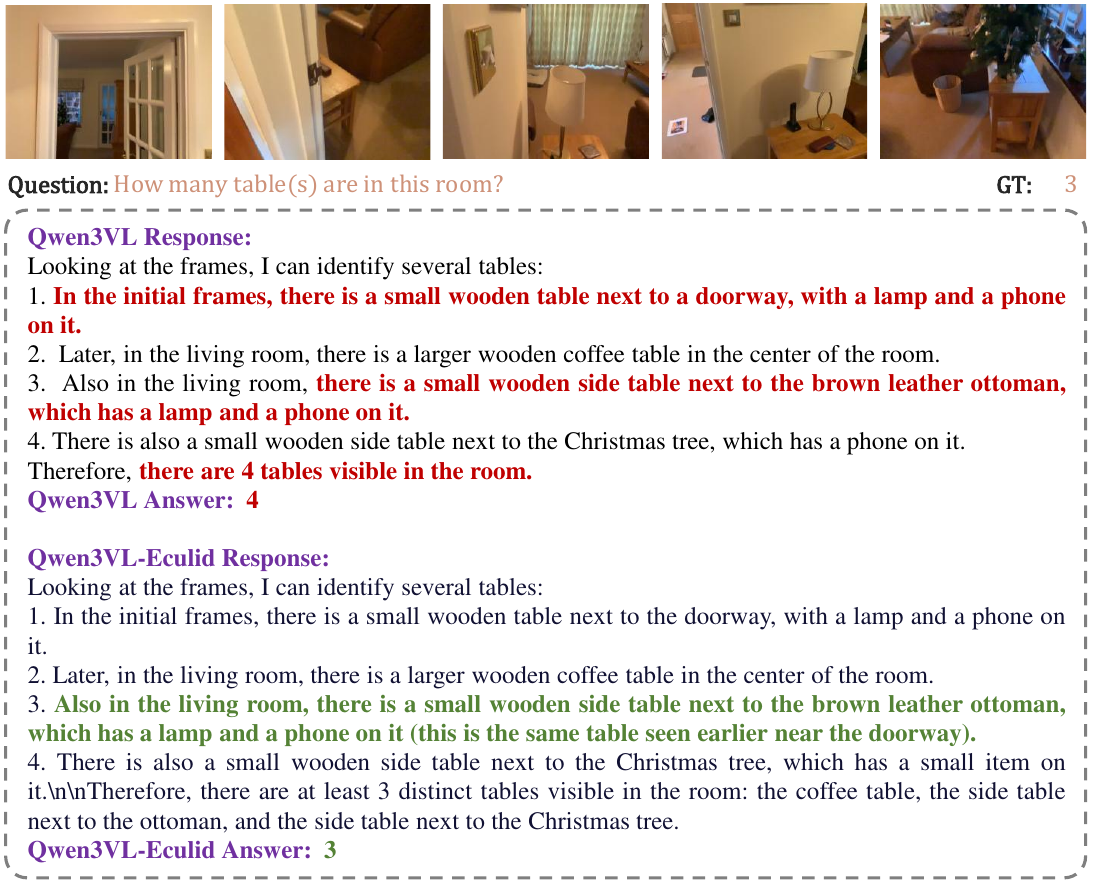}%
    \caption{The response and final answer for Qwen3VL-4B~\cite{Qwen3VL_2025_arXiv} and Qwen3VL-Eculid-4B in VSIBench~\cite{VSIBench_2025_CVPR}.}
    \label{fig:comp3}
\end{figure*}

\begin{figure*}[h]
    \centering
    \includegraphics[width=1\linewidth]{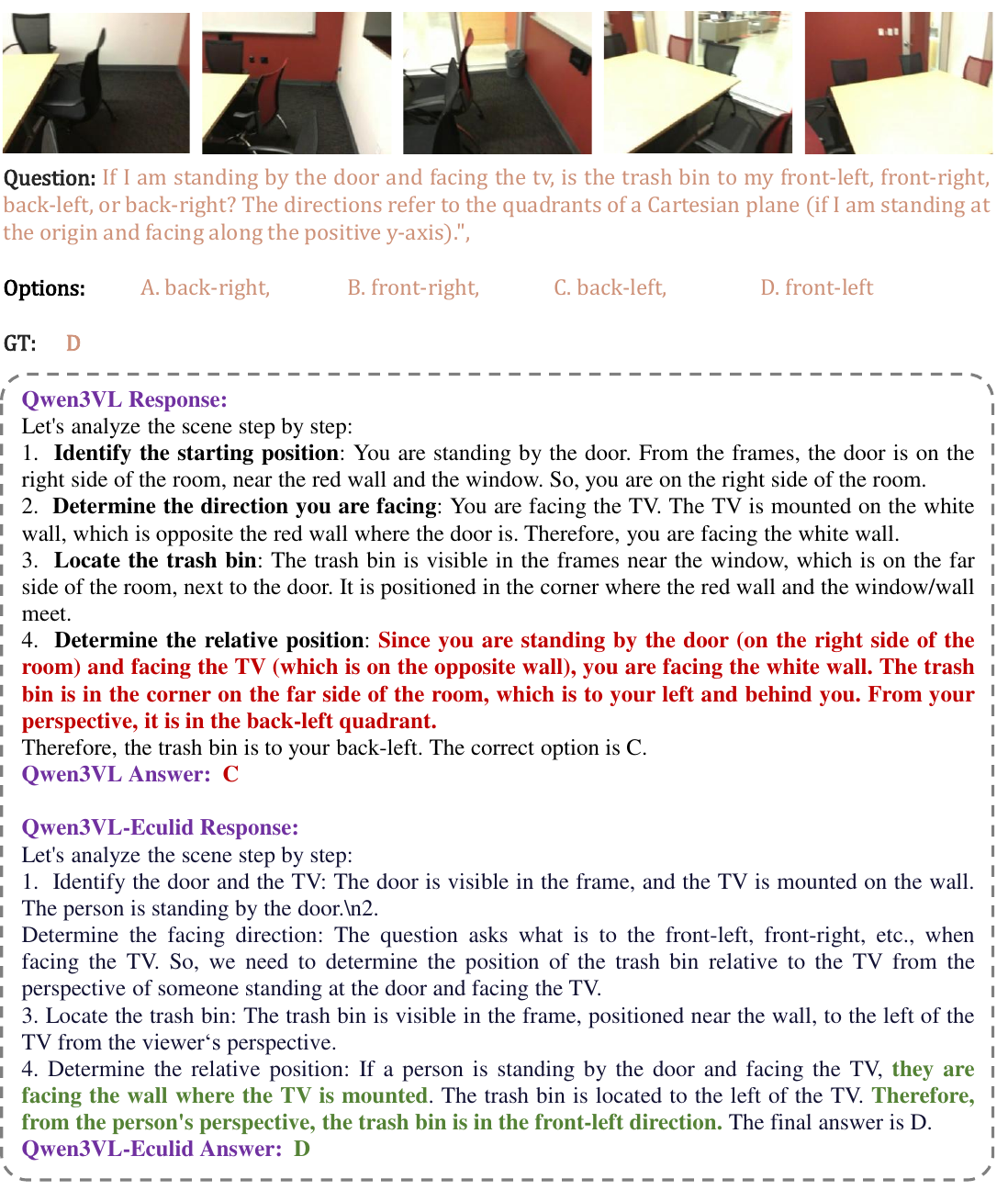}%
    \caption{The response and final answer for Qwen3VL-30B-A3B~\cite{Qwen3VL_2025_arXiv} and Qwen3VL-Eculid-30B-A3B in VSIBench~\cite{VSIBench_2025_CVPR}.}
    \label{fig:comp4}
\end{figure*}

\begin{figure*}[h]
    \centering
    \includegraphics[width=1\linewidth]{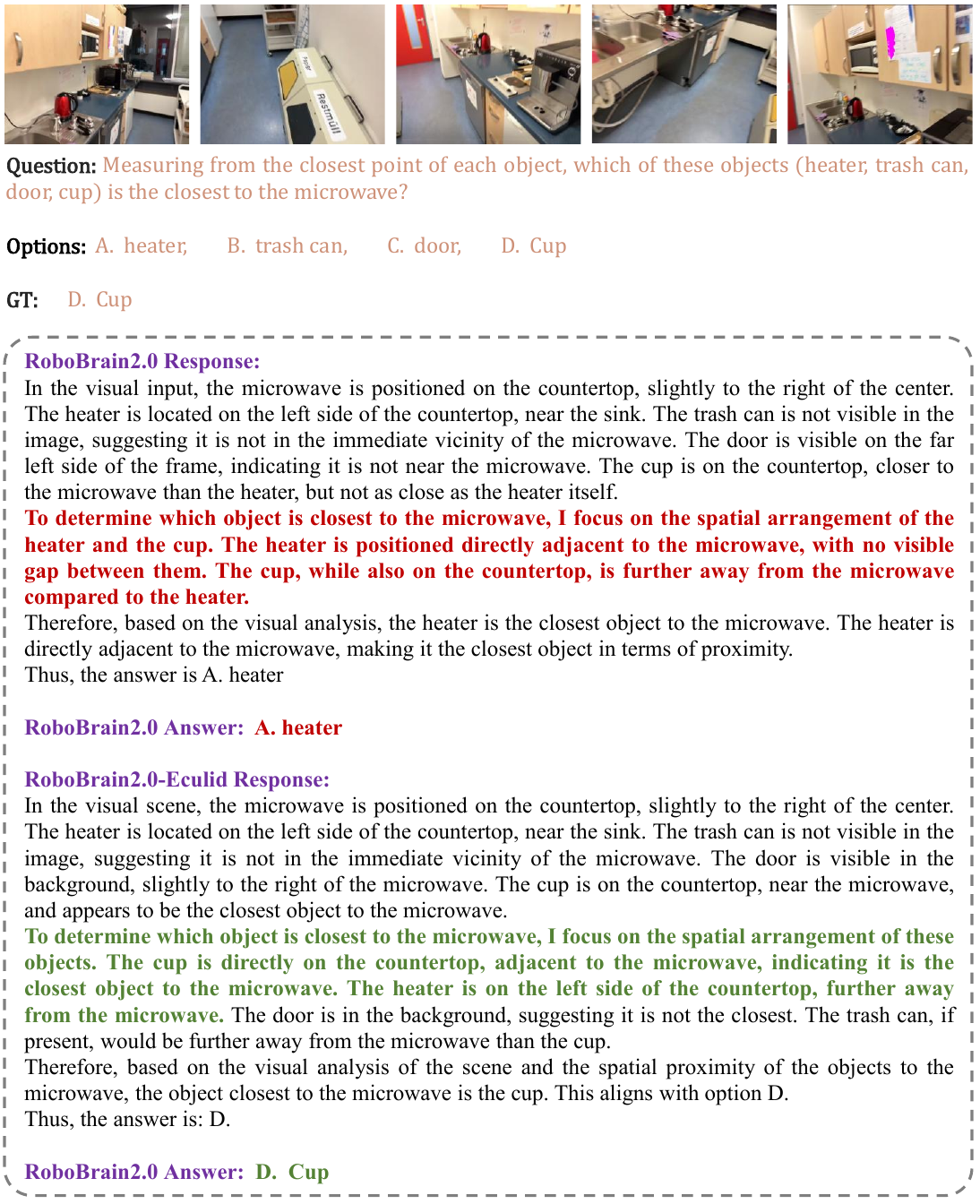}%
    \caption{The response and final answer for RoboBrain2.0-7B~\cite{RoboBrain2.0_2025_TechnicalReport} and RoboBrain2.0-Eculid-7B in VSIBench~\cite{VSIBench_2025_CVPR}.}
    \label{fig:comp5}
\end{figure*}

\begin{figure*}[h]
    \centering
    \includegraphics[width=1\linewidth]{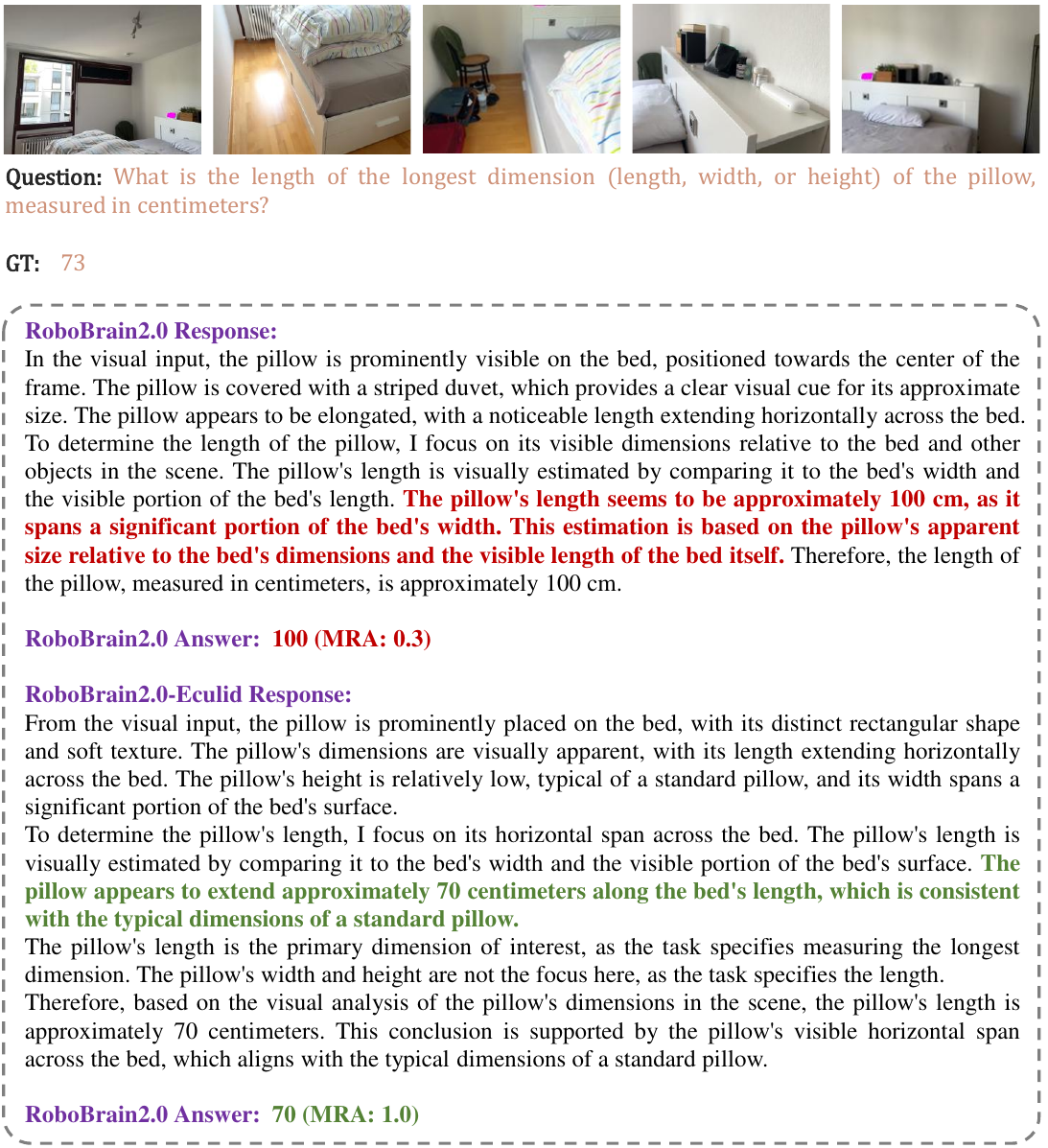}%
    \caption{The response and final answer for RoboBrain2.0-7B~\cite{RoboBrain2.0_2025_TechnicalReport} and RoboBrain2.0-Eculid-7B in VSIBench~\cite{VSIBench_2025_CVPR}.}
    \label{fig:comp6}
\end{figure*}

\begin{figure*}[h]
    \centering
    \includegraphics[width=1\linewidth]{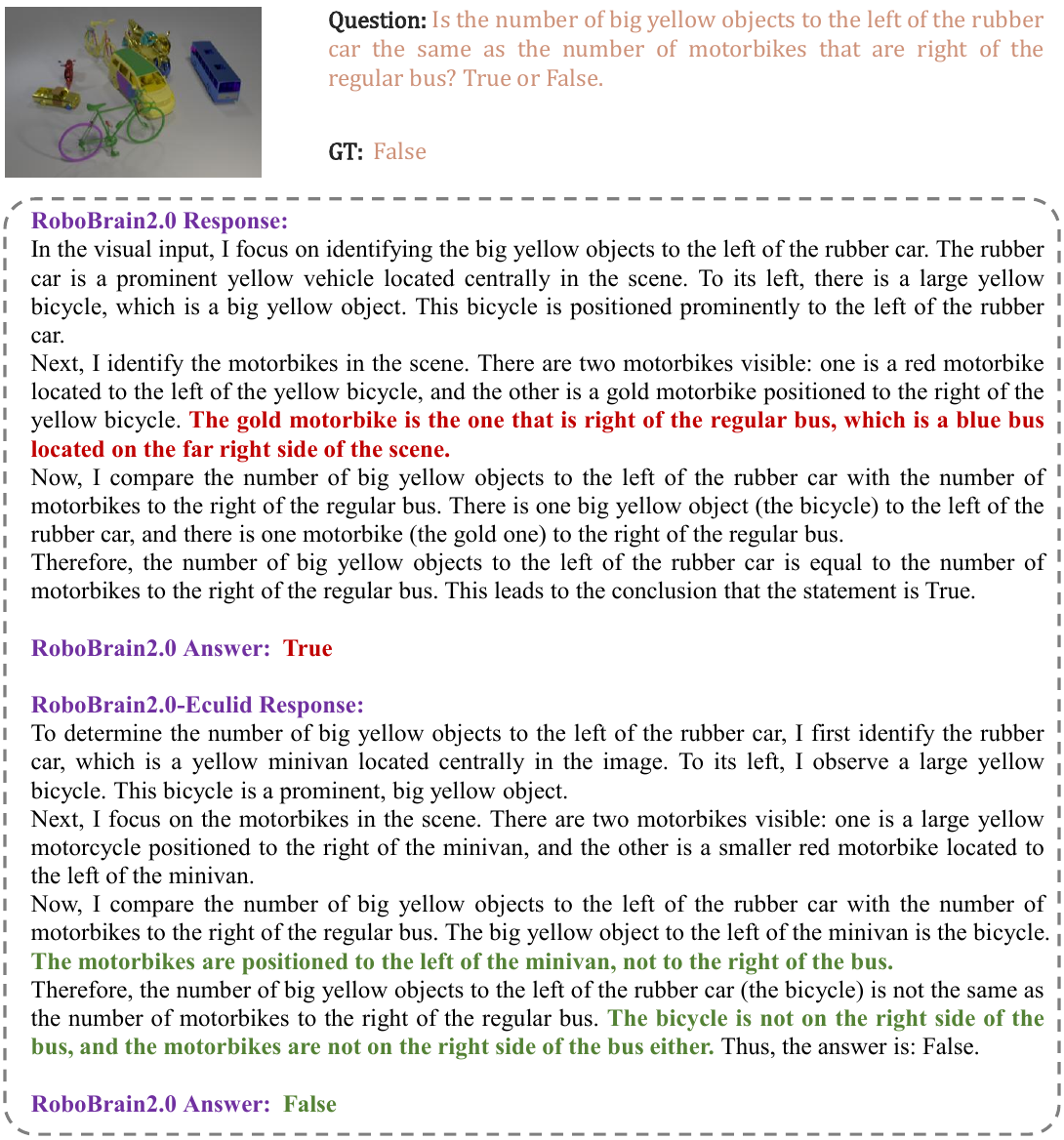}%
    \caption{The response and final answer for RoboBrain2.0-7B~\cite{RoboBrain2.0_2025_TechnicalReport} and RoboBrain2.0-Eculid-7B in SuperCLEVR~\cite{Super-CLEVR_CVPR_2023}.}
    \label{fig:comp7}
\end{figure*}

\begin{figure*}[h]
    \centering
    \includegraphics[width=1\linewidth]{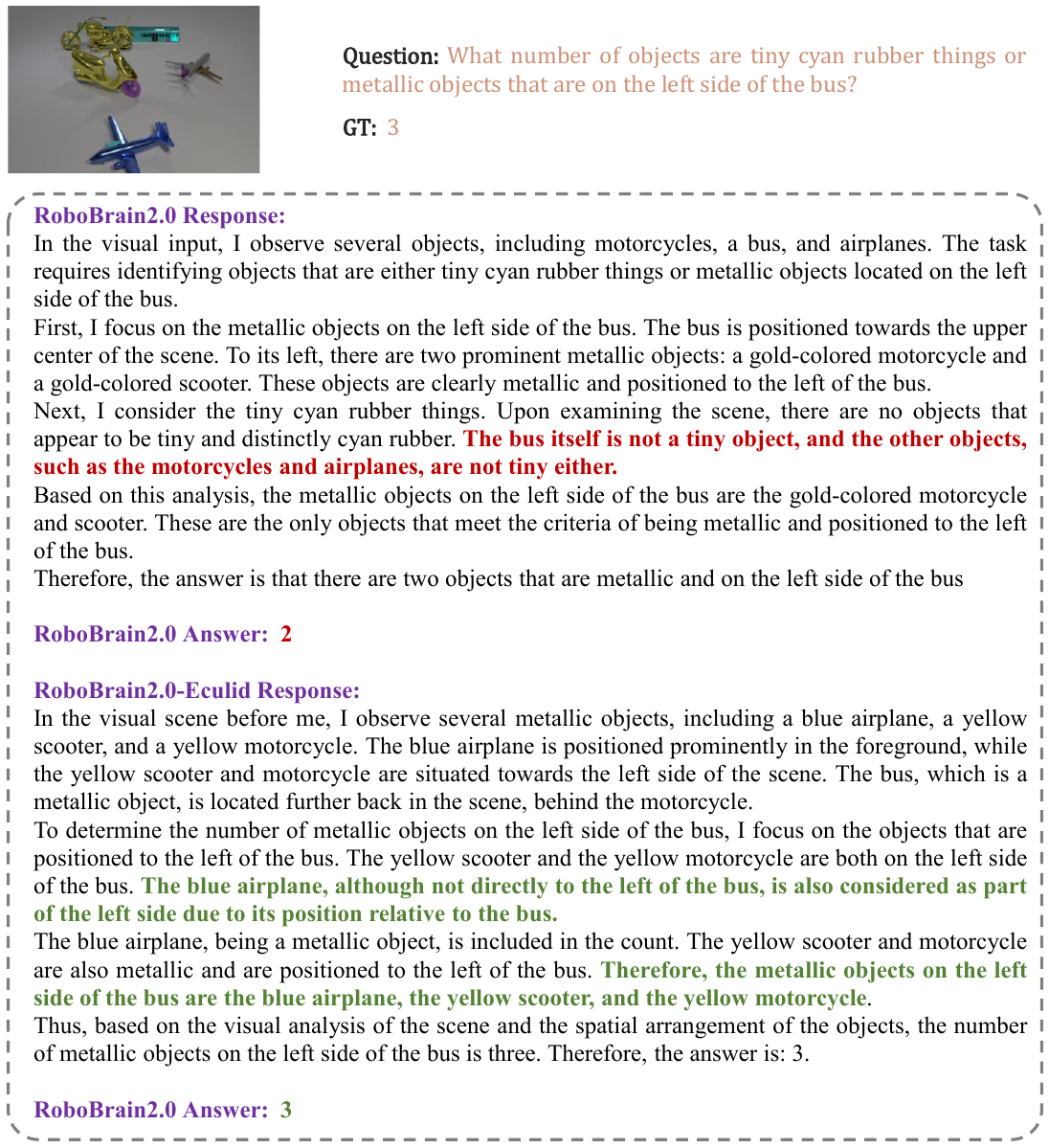}%
    \caption{The response and final answer for RoboBrain2.0-7B~\cite{RoboBrain2.0_2025_TechnicalReport} and RoboBrain2.0-Eculid-7B in SuperCLEVR~\cite{Super-CLEVR_CVPR_2023}.}
    \label{fig:comp8}
\end{figure*}

Beyond the aggregate accuracy gains, Figures\ref{fig:comp1}--\ref{fig:comp8} qualitatively illustrate how geometry tuning alters intermediate reasoning.
After Euclid30K training the model produces more coherent multi view descriptions (Fig.\ref{fig:comp3}), applies geometric similarity relations correctly (Fig.\ref{fig:comp6}), uses quadrant and cardinal directional cues with fewer ambiguities (Fig.\ref{fig:comp4}, Fig.\ref{fig:comp7}), and leverages perspective driven size cues (near–far size scaling) more systematically (Fig.\ref{fig:comp8}).
It also shows clearer distance estimation chains (Fig.\ref{fig:comp1}, Fig.\ref{fig:comp5}), improved object size estimation (Fig.\ref{fig:comp2}, Fig.\ref{fig:comp8}), more reliable counting with cross view consistency (Fig.~\ref{fig:comp3}), and fewer heuristic shortcuts (\textit{e.g.}, premature guesses without spatial justification). 
These qualitative traces are consistent with the quantitative improvements: they suggest the model has internalized foundational Euclidean principles (similarity, proportionality, relative position, viewpoint coherence) and can deploy them across distinct downstream spatial tasks.

\subsection{More experiments about ablation study}\label{ssec:app_ablation}

\cref{tab:appendix_vsibench} presents a detailed analyses of the ablation study results on VSI-Bench~\cite{VSIBench_2025_CVPR}, complementing the summary in \cref{ssec:ablation}.
To isolate the contribution of Euclid30K from potential confounds such as additional training data or GRPO-induced generalization, we trained each model variant on a 30K-sample subset of the spatial-intelligence dataset Clevr-CoGenT~\cite{Clevr_2017_CVPR} using identical hyperparameters, rollout budgets, and training epochs.

\noindent\textbf{Overall Performance}.
Across all model families, Euclid30K training consistently yields higher overall accuracy than Clevr-CoGenT training.
For instance, Qwen2.5VL-3B improves from 29.2\% (base) to 32.0\% with Euclid30K versus 31.3\% with Clevr-CoGenT; Qwen3VL-30B-A3B rises from 40.0\% to 45.8\% with Euclid30K versus 44.2\% with Clevr-CoGenT; and RoboBrain2.0-32B jumps from 43.1\% to 49.6\% with Euclid30K versus 46.7\% with Clevr-CoGenT.
These results indicate that Euclid30K provides a more transferable spatial foundation than an equal volume of non-geometric spatial data.

\input{tab/appendix_puzzle.tex}

\noindent\textbf{Task-Specific Patterns}.
Clevr-CoGenT training yields targeted improvements on tasks closely aligned with its original design, such as object counting and relative direction.
For example, Qwen2.5VL-3B trained on Clevr-CoGenT achieves 40.5\% on object counting, slightly outperforming Euclid30K training (38.3\%), and similar patterns appear for relative direction in several variants.
Conversely, Euclid30K training produces broader gains across multiple categories, particularly on size-estimation tasks (like object size and absolute distance estimation), where geometric reasoning principles directly apply.
Qwen2.5VL-32B trained on Euclid30K reaches 55.8\% on object size versus 44.3\% with Clevr-CoGenT, and 46.0\% on room size versus 40.8\%.

This ablation study reveals a fundamental trade-off: task-specific spatial datasets enhance performance on related categories, whereas structured geometric datasets provide a more general-purpose spatial reasoning foundation that transfers more uniformly across diverse task types.
Euclid30K's strength lies in instilling first-principles geometric knowledge—distance, proportion, similarity, and spatial relations—that applies broadly, even to tasks not explicitly represented in the training corpus.
The pattern holds across Qwen2.5VL, Qwen3VL, and RoboBrain2.0 families, and across parameter scales ranging from 3B to 72B, from dense to MoE.
This consistency suggests that the advantage conferred by geometric surrogate tasks is not an artifact of a particular architecture or capacity regime, but reflects a more general principle: formal Euclidean reasoning offers a robust substrate for spatial intelligence.

In summary, \cref{tab:appendix_vsibench} substantiates the claim that geometry serves as an effective surrogate task for spatial intelligence, with Euclid30K training delivering superior overall performance and broader generalization compared to equal-sized non-geometric spatial datasets.

\subsection{Comparison with Other Surrogate Tasks}

\input{tab/appendix_sft_and_vst.tex}

Beyond our geometry-first curriculum, several studies explore alternative surrogate tasks to strengthen spatial intelligence.
In this subsection, we compare against a representative approach, Spatial-SSRL~\cite{liu2025spatialSSRL}.
Spatial-SSRL proposes a self-supervised pipeline that derives verifiable signals from ordinary RGB/RGB-D images via five pretext tasks capturing 2D/3D spatial structure: shuffled patch reordering, flipped patch recognition, cropped-patch inpainting, regional depth ordering, and relative 3D position prediction.
The released configuration uses 81K image-QA pairs to train a 7B model with RLVR.\@

\cref{tab:appendix_puzzle1} compares a 7B backbone trained on our 30K Euclid30K geometry corpus (RLVR, identical decoding as Sec.~\ref{sec:appendix_exp}) with Spatial-SSRL-7B on VSI-Bench. Euclid30K achieves a higher overall score (31.4 vs. 29.2) with substantially fewer training samples (30K vs. 81K; ~63\% fewer).
Gains concentrate on size- and distance-estimation categories where geometric constraints are directly applicable, while Spatial-SSRL shows stronger performance on route planning and appearance order. Cross-benchmark results in \cref{tab:appendix_puzzle2} show that Euclid30K improves Super-CLEVR and remains competitive on Omni3D-Bench and MindCube under the same inference protocol.

\section{More Discussion}\label{sec:appendix_discussion}

\subsection{Supervised Fine-tuning (SFT).}\label{ssec:sft}
For mathematical problems, collecting step by step solution annotations is substantially more expensive than answer only labels.
Accordingly, Euclid30K does not currently include process annotations, which makes it less suited to SFT routines that rely on explicit reasoning traces.
Nevertheless, geometry QA can still serve as a surrogate task for spatial intelligence under SFT.
In this subsection, in order to show that geometry can as a surrogate task for spatial intelligence not only under reinforcement learning but also under supervised fine tuning, we conduct SFT on Geo170K, a dataset that supplies explicit reasoning trajectories.
Providing these intermediate steps helps the model internalize Euclidean principles.

As shown in \cref{tab:appendix_sft1}, Qwen3VL-30B-A3B improves on VSI Bench from 40.0 to 43.7 overall after Geo170K SFT (the gain is smaller than the 40.0 to 45.8 increase achieved with Euclid30K RL).
In \cref{tab:appendix_sft2}, cross benchmark results also rise on SuperCLEVR (64.1 to 66.5) and Omni3D Bench (36.7 to 40.5), while MindCube shows a modest decrease (39.8 to 38.3).
The decline on MindCube is plausibly due to the composition of Geo170K: most problems are plane geometry items and provide little direct supervision for viewpoint change or three dimensional mental imagery.
Even with this limitation, the consistent gains on the majority of spatial benchmarks reinforce our claim that solving geometry problems is an effective surrogate task for spatial intelligence.

\subsection{Model-Specific Performance Variations}\label{ssec:var}

As noted in Sec.~5.2, we hypothesize that the performance gains after geometry tuning are model specific and depend on capabilities inherited from earlier training data.
In this subsection, we use VST-7B-RL~\cite{VST_arXiv_2025} as a representative case to test this hypothesis.
VST-7B first applies supervised fine-tuning on the VST-P dataset with 4.1M samples that cover 19 spatial skills across single-image, multi-image, and video. It then uses reinforcement learning on the VST-R dataset with 135K reasoning samples.
We start from the released VST-7B-RL checkpoint, add Euclid30K training with GRPO and keep decoding and evaluation identical those to used in the rest of this appendix.
Results in \cref{tab:appendix_vst1} show clear gains on VSI Bench. The overall score moves from 44.3 to 55.5.
\cref{tab:appendix_vst2} shows small improvements on SuperCLEVR and Omni3DBench, and a small decrease on MindCube from 35.5 to 34.8.

A reasonable explanation is that the VST dataset has limited coverage of viewpoint changes and spatiotemporal consistency.
The samples of "camera motion" and "camera-camera" combined account for only about 3\% of the total dataset.
However, MindCube heavily relies on these patterns. 
Meanwhile, the plane geometry problems in Euclid30K do not include explicit supervision signals for viewpoint transformations.
This suggests that the role of Euclid30K is to formalize and reinforce the spatial concepts already present in the model, leading to significant improvements in tasks such as metric estimation and relational reasoning.
However, when certain key concepts (such as robust viewpoint transformation ability) are weakly represented in the pretraining, geometric fine-tuning may offer little benefit and might even slightly interfere with the model's existing heuristic strategies.
These support our hypothesis that models with richer prior spatial concepts obtain larger gains from Euclid30K, while gaps in prior coverage limit or occasionally reduce transfer.

\section{More Visualization about Euclid30K}

\begin{figure*}[h]
    \centering
    \includegraphics[width=0.9\linewidth]{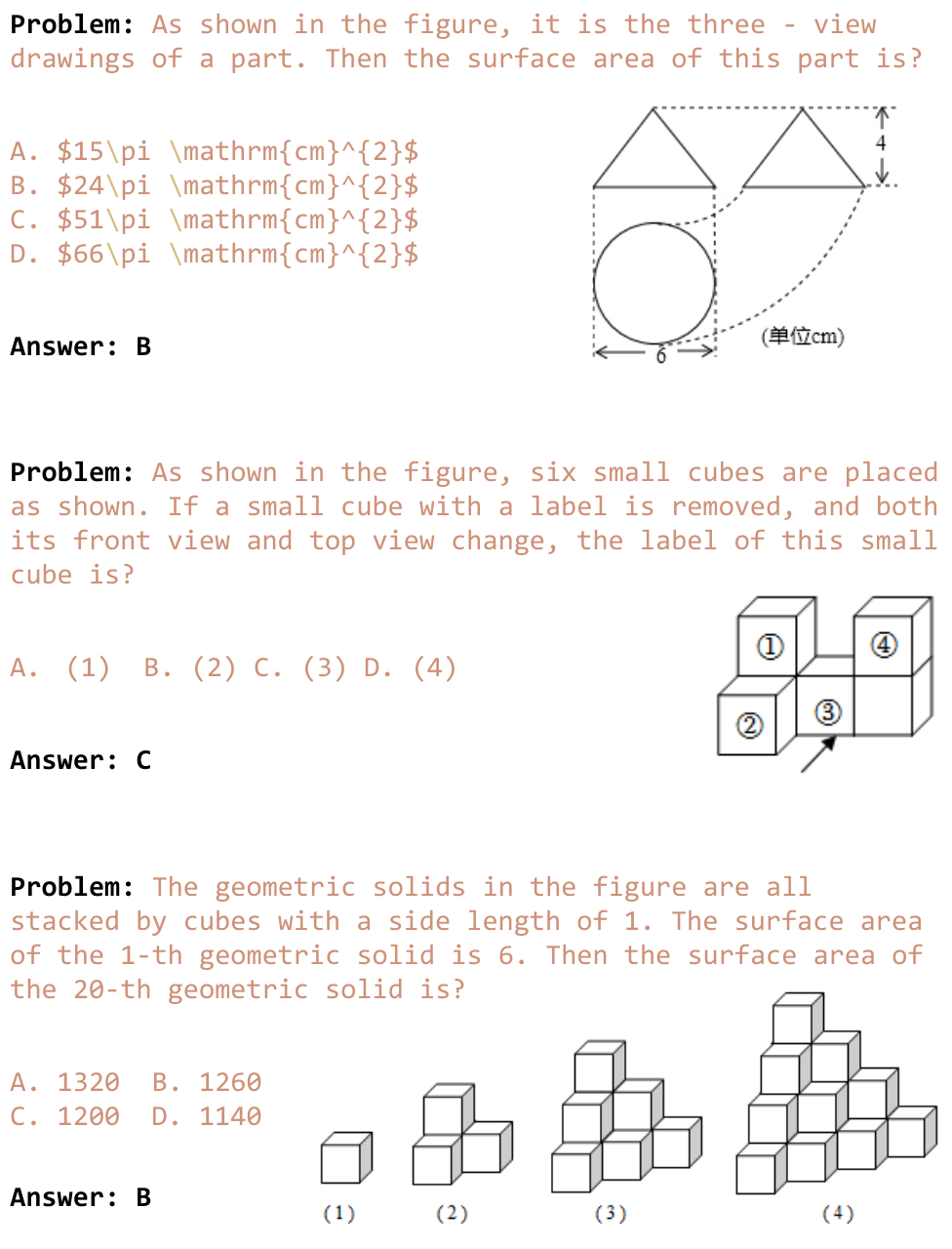}%
    \caption{More examples from the Euclid30K dataset.}
    \label{fig:append_data_show1}
\end{figure*}

\begin{figure*}[h]
    \centering
    \includegraphics[width=0.9\linewidth]{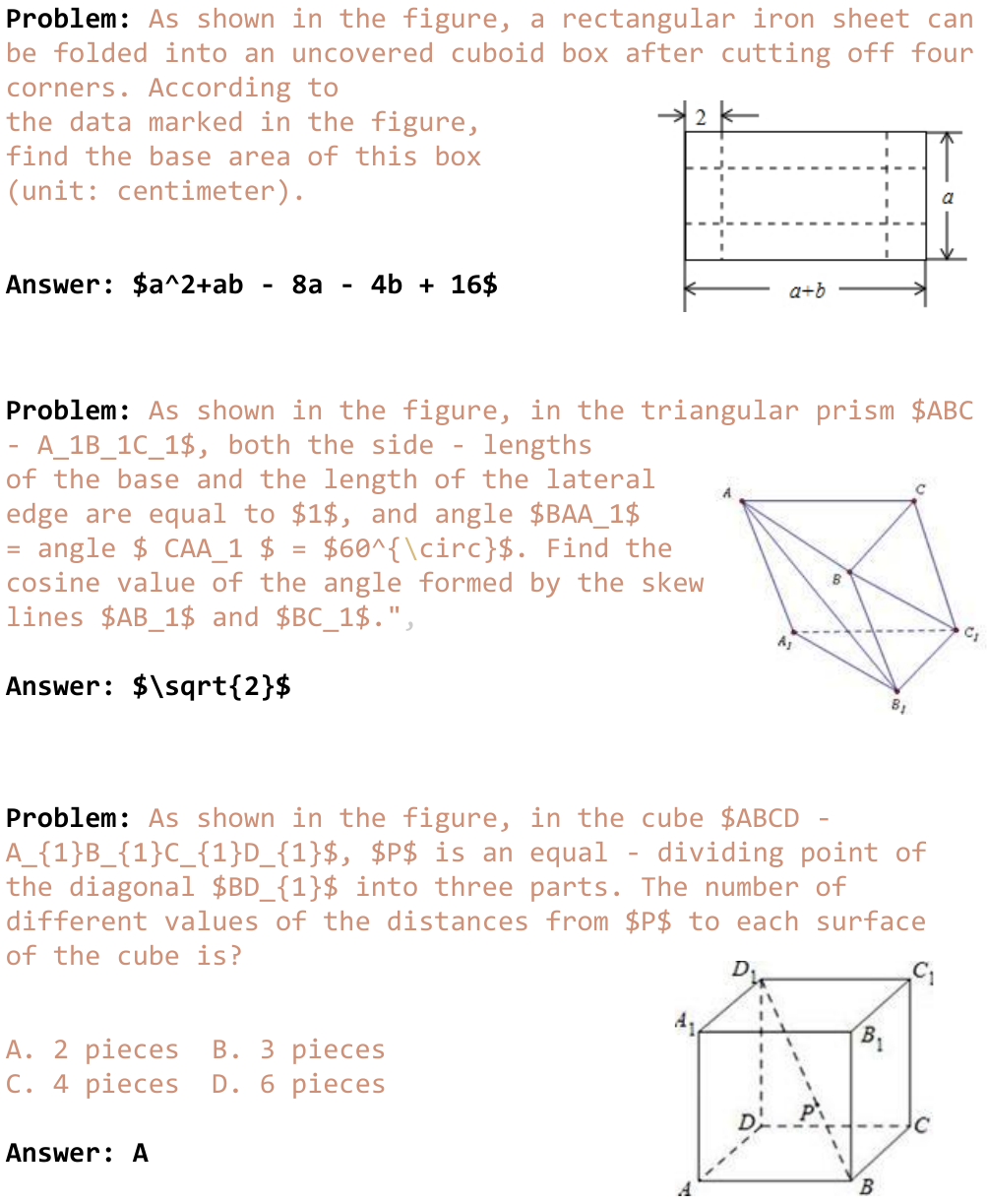}%
    \caption{More examples from the Euclid30K dataset.}
    \label{fig:append_data_show2}
\end{figure*}

\begin{figure*}[h]
    \centering
    \includegraphics[width=0.9\linewidth]{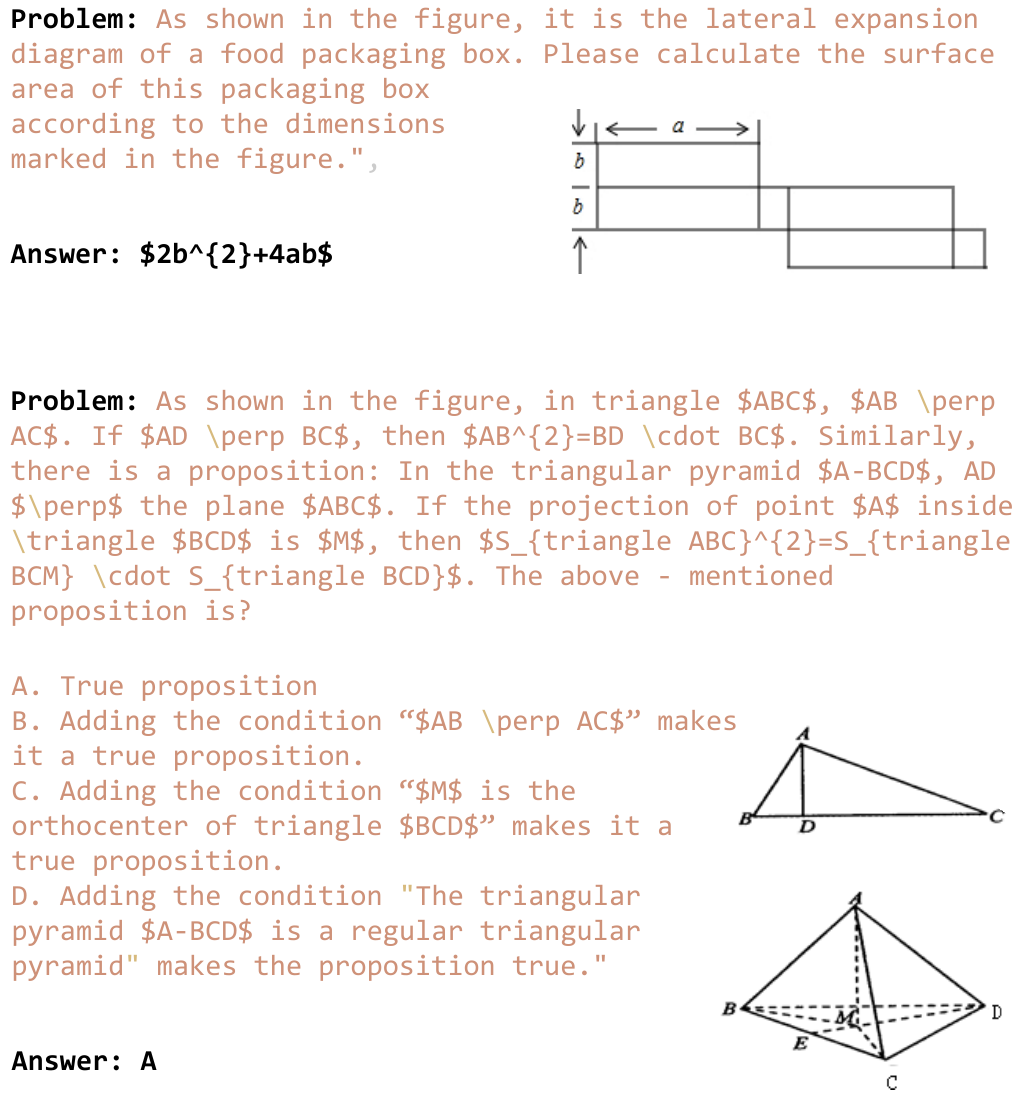}%
    \caption{More examples from the Euclid30K dataset.}
    \label{fig:append_data_show3}
\end{figure*}

Fig.~\ref{fig:append_data_show1}--\ref{fig:append_data_show3} provide additional Euclid30K examples, illustrating diversity in problem types, diagram styles, and reasoning complexity.
These samples complement the main paper by showing the range of geometric configurations and textual formulations present in the Euclid30K dataset.

%% file: tab/appendix_abvsi.tex
\begin{table*}[!t]
\centering
\setlength\tabcolsep{3pt} 
\resizebox{0.97\textwidth}{!}
{
    \begin{tabular}{l|cccc|cccc|c}
        \toprule
        \multirow{2}{*}{\textbf{Methods}}  & \multicolumn{4}{c|}{\textbf{Numerical Question}} & \multicolumn{4}{c|}{\textbf{Multiple-Choice Question}} & \multirow{2}{*}{\textbf{Overall}} \\
        \cmidrule(lr){2-5}\cmidrule(lr){6-9}
         & Obj. Cnt. & Abs. Dist. & Obj. Size & Room Size & Rel. Dist. & Rel. Dir. & Route Plan & Appr. Order &\\
        \midrule
        \rowcolor{navyblue!10}\multicolumn{1}{l|}{\textcolor{black}{\textit{Qwen2.5VL-series}} } & & & & & & & & & \\
         \midrule
        Qwen2.5VL-3B & 35.6 & 23.4 & 34.9 & 16.6 & 34.4 & 40.7 & 26.3 & \textbf{21.8}& 29.2 \\
        Qwen2.5VL-Space-3B  & \textbf{40.5}& 24.5 & 30.1 & \textbf{29.8} & 33.9 & 43.0 & 29.9 & 18.8 & 31.3\\
        Qwen2.5VL-Euclid-3B  &38.3&\textbf{26.8}&\textbf{35.4}&22.2&\textbf{37.0}&\textbf{43.2}&\textbf{36.6}&16.3&\textbf{32.0}\\
         \midrule
        Qwen2.5VL-7B & 39.5 & 17.8 & 16.9 & 5.8 & 33.8 & 36.7 & 24.7 & 22.8 & 24.8 \\
        Qwen2.5VL-Space-7B & \textbf{42.4} &  17.8 &  24.5 &  9.2 &  36.7 &  \textbf{38.5} &  \textbf{29.4} &  23.8 &  27.8 \\
        Qwen2.5VL-Euclid-7B & 38.8 &  \textbf{22.8} &  \textbf{37.3} &  \textbf{23.2} &  \textbf{38.3} &  \textbf{38.5} &  25.8 &  \textbf{26.5} &  \textbf{31.4} \\
        \midrule
        Qwen2.5VL-32B  & 22.4 & 27.0 & 37.9 & 38.0 & 39.4 & \textbf{40.1} & 33.5 & \textbf{41.3} & 34.9 \\
        Qwen2.5VL-Space-32B & \textbf{45.6}& 25.9& 44.3& 40.8& 42.0&39.3& 28.4&35.3& 37.7\\
        Qwen2.5VL-Euclid-32B & 38.7& \textbf{30.9}& \textbf{55.8}& \textbf{46.0}&\textbf{43.7}&37.1& \textbf{34.0}&33.2& \textbf{39.9}\\
        \midrule
        Qwen2.5VL-72B  & 13.6 & 19.6 & 40.9 & 41.1 & 37.7 & 35.3 & \textbf{34.0} & 36.2 & 32.3 \\
        Qwen2.5VL-Space-72B & 15.6 & 24.8 & 40.7 & 41.4 & 43.4 & \textbf{37.8}& 29.4 & 33.5 & 33.2\\
        Qwen2.5VL-Euclid-72B  &  \textbf{22.5} & \textbf{27.2} & \textbf{55.7} &  \textbf{43.3} & \textbf{44.9} &  37.1 &  32.5 &  \textbf{36.6} & \textbf{37.5}\\
        \midrule
        \rowcolor{navyblue!10}\multicolumn{1}{l|}{\textcolor{black}{\textit{Qwen3VL-series}}} & & & & & & & & & \\
        \midrule
        Qwen3VL-4B & 28.5 & 33.0 & 32.6 & 43.5 & 40.3 & 40.0 & 33.0 & 33.2 & 35.5 \\
        Qwen3VL-Space-4B & 32.1 & 37.1 & 44.7 & \textbf{51.0} & \textbf{49.3} & 43.8 & \textbf{37.6} & 38.5 & 41.8 \\
        Qwen3VL-Euclid-4B & \textbf{33.3} & \textbf{37.4} & \textbf{49.5} & 48.3 & 46.5 & \textbf{46.3} & 34.0 & \textbf{42.9} & \textbf{42.3} \\
        \midrule
        Qwen3VL-30B-A3B & 27.4 & 32.3 & 53.6 & \textbf{44.0} & 42.1 & 36.2 & 35.5 & 48.9 & 40.0 \\
        Qwen3VL-Space-30B-A3B & 29.9 & 36.8 & 58.0 & 43.4 & 47.9 & 49.1 & \textbf{35.9} & 52.7 & 44.2\\
        Qwen3VL-Euclid-30B-A3B & \textbf{33.5} & \textbf{37.5} & \textbf{64.1} & 43.1 & \textbf{48.7} & \textbf{49.6} & 35.6 & \textbf{54.4} & \textbf{45.8} \\
        \midrule
        \rowcolor{navyblue!10}\multicolumn{1}{l|}{\textcolor{black}{\textit{RoboBrain2.0-series}}} & & & & & & & & & \\
        \midrule
        RoboBrain2.0-7B& 46.0 & 32.7 & 58.9 & 35.9 & 45.9 & 41.5 & 30.9 & 55.2 & 43.0 \\
        RoboBrain2.0-Space-7B& \textbf{66.4}& 34.4& 65.8& \textbf{41.0}& 46.6& \textbf{46.5}& \textbf{36.6}& 52.3& 48.7\\
        RoboBrain2.0-Euclid-7B& \textbf{66.4}& \textbf{36.9}& \textbf{66.3}& 40.5& \textbf{48.3}& 45.3& 35.6& \textbf{57.8}& \textbf{49.6}\\
        \midrule
        RoboBrain2.0-32B& 50.5 & 37.0 & 59.2 & 28.4 & 43.2 & 46.1 & \textbf{34.5}& 39.5 & 43.1 \\
        RoboBrain2.0-Space-32B& 58.0 & 36.9 & 62.2 &\textbf{47.8} & 46.9 & 44.5& 34.0& 42.1 & 46.7\\
        RoboBrain2.0-Euclid-32B& \textbf{59.2} &  \textbf{39.4} & \textbf{63.4} & \textbf{47.8}& \textbf{48.7} & \textbf{47.5} & 33.5 & \textbf{57.0} & \textbf{49.6} \\
        \bottomrule
    \end{tabular}
}
\captionof{table}{\small \textbf{Ablation experiment on VSI-Bench}~\cite{VSIBench_2025_CVPR}. We compare training a model on a 30K subset of the spatial intelligence dataset Clevr-CoGenT v.s. the geometric dataset Euclid30K to verify that the geometric dataset serves as a surrogate task to improve the spatial intelligence capabilities of the model. \textbf{Bolding} indicates the best score within each model type.}
\label{tab:appendix_vsibench}
\end{table*}

%% file: tab/appendix_puzzle.tex
\begin{table*}[!t]
\centering
\setlength\tabcolsep{3pt} 
\resizebox{0.97\textwidth}{!}
{
    \begin{tabular}{l|cccc|cccc|c}
        \toprule
        \multirow{2}{*}{\textbf{Methods}}  & \multicolumn{4}{c|}{\textbf{Numerical Question}} & \multicolumn{4}{c|}{\textbf{Multiple-Choice Question}} & \multirow{2}{*}{\textbf{Overall}} \\
        \cmidrule(lr){2-5}\cmidrule(lr){6-9}
         & Obj. Cnt. & Abs. Dist. & Obj. Size & Room Size & Rel. Dist. & Rel. Dir. & Route Plan & Appr. Order &\\
        \midrule
        Qwen2.5VL-7B & \textbf{39.5} & 17.8 & 16.9 & 5.8 & 33.8 & 36.7 & 24.7 & 22.8 & 24.8 \\
        Spat1-SSRL-7B \cite{liu2025spatialSSRL} & 37.1 & \textbf{24.6} & 22.9 & 17.0 & 36.3 & 35.1 &\textbf{33.5} & \textbf{27.3} & 29.2 \\
        Qwen2.5VL-Euclid-7B & 38.8 &  22.8 &  \textbf{37.3} &  \textbf{23.2} &  \textbf{38.3} &  \textbf{38.5} &  25.8 &  26.5 &  \textbf{31.4} \\
         \bottomrule
    \end{tabular}
}
\captionof{table}{\small \textbf{Comparisons with Spat1-SSRL-7B on VSI-Bench}. \textbf{Bolding} indicates the best score within each model type.}
\vspace{-0.5em}\label{tab:appendix_puzzle1}
\end{table*}

\begin{table}[!t]
\centering
\renewcommand{\arraystretch}{1.2}
\setlength\tabcolsep{3pt} 
\resizebox{0.94\linewidth}{!}
{
  \begin{tabular}{l|ccc}
    \toprule
    \textbf{Methods}   & \textbf{SuperClevr} &  \textbf{Omni3DBench} & \textbf{MindCube} \\
    \midrule
     Qwen2.5VL-7B  & 76.1 & 28.3 & 30.0   \\
     Spat1-SSRL-7B \cite{liu2025spatialSSRL} & 76.3 & \textbf{33.1} & 30.6 \\
     Qwen2.5VL-Euclid-7B & \textbf{86.2} & 31.1 & \textbf{31.1}\\
    \bottomrule
  \end{tabular}
}
\captionof{table}{\small \textbf{Comparisons with Spat1-SSRL-7B on SuperClevr, Omni3D Bench, and MindCube.} \textbf{Bolding} indicates the best score within each model type.}\label{tab:appendix_puzzle2}
\vspace{-3mm}
\end{table}

%% file: tab/appendix_sft_and_vst.tex
\begin{table*}[!t]
\centering
\setlength\tabcolsep{3pt} 
\resizebox{0.97\textwidth}{!}
{
    \begin{tabular}{l|cccc|cccc|c}
        \toprule
        \multirow{2}{*}{\textbf{Methods}}  & \multicolumn{4}{c|}{\textbf{Numerical Question}} & \multicolumn{4}{c|}{\textbf{Multiple-Choice Question}} & \multirow{2}{*}{\textbf{Overall}} \\
        \cmidrule(lr){2-5}\cmidrule(lr){6-9}
         & Obj. Cnt. & Abs. Dist. & Obj. Size & Room Size & Rel. Dist. & Rel. Dir. & Route Plan & Appr. Order &\\
        \midrule
        Qwen3VL-30B-A3B & 27.4 & 32.3 & 53.6 & \textbf{44.0} & 42.1 & 36.2 & \textbf{35.5} & 48.9 & 40.0 \\
        + GeometricSFT & \textbf{29.6} & \textbf{33.8} & \textbf{56.9} & 39.1 & \textbf{46.8} & \textbf{47.8} & 35.1 & \textbf{60.2} & \textbf{43.7} \\
         \bottomrule
    \end{tabular}
}
\captionof{table}{\small \textbf{Evaluation on VSI-Bench}. + GeometricSFT indicate the Qwen3VL-30B-A3B trained with SFT on the Geo170K dataset \cite{GEO170K_2025_ICLR}. \textbf{Bolding} indicates the best score within each model type.}
\label{tab:appendix_sft1}
\end{table*}

\begin{table*}[!t]
\centering
\setlength\tabcolsep{3pt} 
\resizebox{0.97\textwidth}{!}
{
    \begin{tabular}{l|cccc|cccc|c}
        \toprule
        \multirow{2}{*}{\textbf{Methods}}  & \multicolumn{4}{c|}{\textbf{Numerical Question}} & \multicolumn{4}{c|}{\textbf{Multiple-Choice Question}} & \multirow{2}{*}{\textbf{Overall}} \\
        \cmidrule(lr){2-5}\cmidrule(lr){6-9}
         & Obj. Cnt. & Abs. Dist. & Obj. Size & Room Size & Rel. Dist. & Rel. Dir. & Route Plan & Appr. Order &\\
        \midrule
        Qwen2.5VL-7B & 39.5 & 17.8 & 16.9 & 5.8 & 33.8 & 36.7 & 24.7 & 22.8 & 24.8 \\
        VST-7B~\cite{VST_arXiv_2025} & 56.5 & 35.1 & 69.0 & 53.3 & 50.1 & 11.7 & 28.4 & 50.3 & 44.3 \\
        VST-Euclid-7B & \textbf{66.4} & \textbf{38.4} & \textbf{74.2} & \textbf{60.1} & \textbf{56.5} & \textbf{48.6} & \textbf{41.8} & \textbf{57.8} & \textbf{55.5} \\
         \bottomrule
    \end{tabular}
}
\captionof{table}{\small \textbf{Evaluation on VSI-Bench }. VST-Euclid indicate the VST~\cite{VST_arXiv_2025} trained with GRPO~\cite{GRPO_DeepSeekMath_arXiv_2024} on the Euclid30K dataset. \textbf{Bolding} indicates the best score within each model type.}\label{tab:appendix_vst1}
\end{table*}

\begin{table}[!t]
\centering
\renewcommand{\arraystretch}{1.2}
\setlength\tabcolsep{3pt} 
\resizebox{0.94\linewidth}{!}
{
  \begin{tabular}{l|ccc}
    \toprule
    \textbf{Methods}   & \textbf{SuperClevr} &  \textbf{Omni3DBench} & \textbf{MindCube} \\
    \midrule
    Qwen3VL-30B-A3B  & 64.1 & 36.7 & \textbf{39.8} \\
    + GeometricSFT  &\textbf{66.5} & \textbf{40.5} & 38.3 \\
    \bottomrule
  \end{tabular}
}
\captionof{table}{\small \textbf{Evaluation on SuperClevr, Omni3D Bench, and MindCube.} + GeometricSFT indicate the Qwen3VL-30B-A3B trained with SFT on the Geo170K dataset \cite{GEO170K_2025_ICLR}. \textbf{Bolding} indicates the best score within each model type.}
\label{tab:appendix_sft2}
\end{table}

\begin{table}[!t]
\centering
\renewcommand{\arraystretch}{1.2}zq
\setlength\tabcolsep{3pt} 
\resizebox{0.94\linewidth}{!}
{
  \begin{tabular}{l|ccc}
    \toprule
    \textbf{Methods}   & \textbf{SuperClevr} &  \textbf{Omni3DBench} & \textbf{MindCube} \\
    \midrule
     Qwen2.5VL-7B  & 76.1 & 28.3 & 30.0   \\
     VST-7B~\cite{VST_arXiv_2025} & 83.1 & 36.5 & \textbf{35.5} \\
     VST-Euclid-7B & \textbf{86.3} & \textbf{37.1} & 34.8\\
    \bottomrule
  \end{tabular}
}
\captionof{table}{\small \textbf{Comparisons with Spat1-SSRL-7B on SuperClevr, Omni3D Bench, and MindCube.} VST-Euclid indicate the VST~\cite{VST_arXiv_2025} trained with GRPO~\cite{GRPO_DeepSeekMath_arXiv_2024} on the Euclid30K dataset. \textbf{Bolding} indicates the best score within each model type.}\label{tab:appendix_vst2}
\end{table}

\begin{table}[t]
\centering
\renewcommand{\arraystretch}{1.05}
\setlength{\tabcolsep}{8pt}
\begin{tabular}{c|ll}
\toprule
\textbf{Category} & \textbf{Parameter} & \textbf{Value} \\
\midrule
Dataset & image max pixels & 262144 \\
Dataset & video max pixels & 16384 \\
Dataset & cutoff len & 8192 \\
\midrule
Method & finetuning type & lora \\
Method & lora rank & 8 \\
Method & lora target & all \\
\midrule
Train & per device train batch size & 4 \\
Train & gradient accumulation steps & 4 \\
Train & learning rate & 1e-4 \\
Train & num train epochs & 1 \\
Train & lr scheduler type & cosine \\
Train & weight decay & 0.01 \\
Train & warmup ratio & 0.1 \\
Train & bf16 & true \\
\bottomrule
\end{tabular}\caption{\small \textbf{Supervised Fine-tuning (SFT) configuration in LLaMAFactory framework \cite{llamafactory_2024_acl}.}}\label{tab:sft_config}
\vspace{-5mm}
\end{table}